\documentclass[preprint,review,number]{elsarticle}

\usepackage{amssymb}
\usepackage{booktabs}
\usepackage{tcolorbox}
\usepackage{verbatim}
\usepackage{listings}
\usepackage{xcolor}
\usepackage{graphicx}
\usepackage{subfig}
\usepackage{tabularx}
\usepackage{placeins}
\usepackage{url}
\usepackage{todonotes}
\usepackage[numbers]{natbib}
\usepackage{amsmath}

\journal{Journal of Information Security and Applications}

\begin{document}

\begin{frontmatter}


\title{LLMs for Domain Generation Algorithm Detection}

\author[inst1,inst3]{Reynier Leyva La O\corref{cor1}}%\thanks{Corresponding author}}
\ead{rleyvalao@mendoza-conicet.gob.ar}

\author[inst2,inst3]{Carlos A. Catania}
\ead{harpo@ingenieria.uncuyo.edu.ar}

\author[inst2,inst3]{Tatiana S. Parlanti}
\ead{tatiana.parlanti@ingenieria.uncuyo.edu.ar}

%\author[inst1]{Rodrigo Gonzalez}

\affiliation[inst1]{organization={GridTICs, Facultad Regional Mendoza, Universidad Tecnológica Nacional},%Department and Organization
            addressline={Rodriguez 273}, 
            %city={Mendoza},
            postcode={M5502AJE}, 
            state={Mendoza},
            country={Argentina}}

\affiliation[inst2]{organization={Facultad de Ingeniería, Universidad Nacional de Cuyo},%Department and Organization
            addressline={Centro Universitario}, 
            %city={Mendoza},
            postcode={M5502JMA}, 
            state={Mendoza},
            country={Argentina}}
            
\affiliation[inst3]{organization={National Scientific and Technical Research Council (CONICET)},%Department and Organization
            addressline={Godoy Cruz 2290}, 
            %city={CABA},
            postcode={C1425FQB}, 
            state={CABA},
            country={Argentina}}

\cortext[cor1]{Corresponding author}
            
\begin{abstract}

This work analyzes the use of large language models (LLMs) for detecting domain generation algorithms (DGAs). We perform a detailed evaluation of two important techniques: In-Context Learning (ICL) and Supervised Fine-Tuning (SFT), showing how they can improve detection. SFT increases performance by using domain-specific data, whereas ICL helps the detection model to quickly adapt to new threats without requiring much retraining. We use Meta’s Llama3 8B model, on a custom dataset with 68 malware families and normal domains, covering several hard-to-detect schemes, including recent word-based DGAs. Results proved that LLM-based methods can achieve competitive results in DGA detection. In particular, the SFT-based LLM DGA detector outperforms state-of-the-art models using attention layers, achieving 94\% accuracy with a 4\% false positive rate (FPR) and excelling at detecting word-based DGA domains.
\end{abstract}


\end{frontmatter}

%\linenumbers

\section{Introduction}

%\subsection{The Significance of DGA Detection in Cybersecurity}

In the ever-evolving landscape of cybersecurity, domain generation algorithms (DGAs) have emerged as a significant threat, posing unique challenges to traditional security measures. DGAs are sophisticated tools employed by cybercriminals to dynamically create large numbers of domain names, primarily used to establish and maintain command and control (C\&C) infrastructure for botnets and other malicious activities. 

The operation of a DGA is based on an initial seed or key, from which sequences of characters are generated to form domain names. These sequences can vary in length and structure, and encryption and obfuscation techniques are often used to further hinder their detection.

DGAs pose significant challenges to cybersecurity by enabling botnets to frequently change command-and-control servers, making them resilient to takedown efforts and difficult for defenders to block. They also facilitate covert data exfiltration, malware distribution, and phishing campaigns by generating constantly shifting domains that evade detection by traditional security measures. Additionally, DGAs bypass reputation-based systems with newly created domains lacking history, overwhelming security tools with high volumes of domains, increasing false positives, and straining network resources.

Given these challenges, effective DGA detection has become a critical component of modern cybersecurity strategies. Traditional approaches, such as static blacklists and simple heuristics, have proven inadequate against the dynamic nature of DGAs. This has led to increased interest in more sophisticated detection methods, including machine learning and deep learning \cite{saeed2021survey, li2019domain, zhou2019cnn, hwang2020effective} as they have shown promise in enhancing DGA detection by leveraging more sophisticated feature extraction and classification methods. Similarly, and more recently, the application of large language models (LLMs) has also attracted interest \cite{dombert}.

In this context, LLMs, thanks to their extensive training data and ability to comprehend semantic patterns, offer new potential for effective DGA detection \cite{llama3_github, llama3_huggingface}. LLMs' adaptability and semantic understanding are crucial for analyzing the complex patterns generated by DGAs. Furthermore, LLMs do not require extensive datasets for their application, making them efficient for deployment in dynamic environments.

%However, LLMs also present significant challenges, such as high computational power requirements and slower processing times compared to traditional models. Additionally, the black-box nature of cloud-based LLMs poses security concerns, making them less suitable for applications requiring transparency and explainability.

%\subsection{Proposed Approach and Paper Structure}

%Despite the above, this work explores the application of LLMs for DGA detection. Although LLMs exhibit longer processing times compared to methods such as convolutional neural networks (CNNs), they are proposed for use in a novel layered detection approach due to their superior accuracy and lower false positive rates (FPR). In this strategy, LLMs are positioned as a secondary verification layer behind faster, simpler detectors. This approach is particularly beneficial for applications requiring low FPR, a criterion often challenging for rapid models to meet. Utilizing LLMs in this manner allows for the harnessing of their precision and reliability to significantly enhance the overall effectiveness of DGA detection systems without compromising speed.

In this work, we analyze the performance of LLM-based methods for DGA detection, comparing them with state-of-the-art models. Our focus is on recent local models, such as Meta's Llama 3 8B. We explore two main strategies: In-Context Learning (ICL) and Supervised Fine-Tuning (SFT). The evaluation is conducted on a carefully designed dataset that includes examples from 68 distinct DGA families, covering various schemes, including recent word-based DGAs. These word-based schemes generate domains by concatenating sequences of words from one or more wordlists, resulting in domains that appear less random and are therefore more challenging to detect \cite{plohmann2016comprehensive}.  

Our results indicate that SFT with domain-specific data significantly improves detection capabilities in particular reducing the false positive rate (FPR), whereas ICL enables rapid adaptation to new and evolving threats without extensive retraining. This research highlights the potential of LLMs to enhance cybersecurity defenses against DGA-based attacks, providing a comprehensive solution that balances speed and accuracy.

The main contributions of the paper are:
\begin{itemize}
    \item A dataset with 68 malware families and normal domains from the Tranco dataset~\cite{tranco}.
    \item A complete analysis of  the potential of ICL and SFT for improving DGA detection using LLMs.
    \item A state-of-the-art LLM-based DGA detector with lower FPR and better detection of DGA word-based domains.
\end{itemize}

The remainder of this paper is organized as follows: Section \ref{sec:background} provides technical background on LLMs for DGA detection. Section \ref{sec:Previos Work} reviews previous works in DGA detection. Section \ref{sec:methodology} describes the proposed approach, including the dataset and methodology. Section \ref{sec:Experiment} outlines the design and results of experiments, focusing on ICL and fine-tuning approaches. Section \ref{sec:Discution} discusses the practical implications of the findings and proposes a strategy for practical implementation. Finally, Section \ref{sec:conclusions} concludes the paper and suggests directions for future research.

\section{Background}
\label{sec:background}

To better understand how large language models (LLMs) can be applied to DGA detection, it is critical to explore in detail the technical aspects of these models and the specific approaches used in this study.

\subsection{Large Language Models}

LLMs are based on the Transformer architecture, which fundamentally relies on self-attention mechanisms to process and generate sequential data \cite{devlin2019bert, brown2020language, vaswani2017attention}. This architecture allows the model to weigh the relevance of different input parts dynamically, leading to a more nuanced understanding of context and relationships within the data. Pre-trained on extensive amounts of text data, these models can learn intricate patterns and linguistic structures \cite{liu2019roberta, wei2022chain}.

Transformers are composed of several key components. The embedding layer converts input tokens into dense vector representations, enabling the model to handle various types of data inputs uniformly \cite{brown2020language}. The multi-head attention mechanism is crucial as it allows the model to focus on different parts of the input simultaneously, capturing complex dependencies that might exist within the data. This is complemented by feedforward neural networks, which process the output of the attention layers and introduce non-linearity, enhancing the model's ability to understand and predict complex patterns. Additionally, layer normalization is applied throughout to stabilize and accelerate the learning process, ensuring consistent model performance.

%TP- Specifically, the Llama3 8B model consists of 8 billion parameters spread across multiple transformer layers. This extensive parameterization enables the model to encapsulate and recognize convoluted patterns in data, showcasing the transformative power of LLMs in various applications \cite{llama3_github, llama3_huggingface, huang2024good}.  

%\todo[inline] {add a paragraph mentioning our interpretation of training. What is training for ICL and finetuned}

\subsection{Model}
Llama3 with 8 Billion parameters spread across multiple transformer layers, as known as \textit{Llama3 8B model},  was used in this study. Its extensive parameterization enables it to encapsulate and recognize convoluted patterns in data  \cite{llama3_github, llama3_huggingface, huang2024good}. 

Llama3 is an auto-regressive language model that uses an optimized transformer architecture. It uses a tokenizer with a vocabulary of 128K tokens and was trained on sequences of 8,192 tokens. Grouped-query attention (GQA) is used for all models to improve inference efficiency. The tuned versions use supervised fine-tuning (SFT) and reinforcement learning with human feedback (RLHF) to align with human preferences for helpfulness and safety.

\subsection{In-Context Learning}
In-context learning (ICL) represents a novel approach where LLMs adapt to new tasks by leveraging their pre-existing knowledge without requiring much retraining \cite{brown2020language, agarwal2024many, lu2021fantastically}. This capability is particularly powerful as it allows for rapid adaptation to a wide range of tasks by simply providing examples and context within the model's input window. The process begins with prompt engineering, where a carefully crafted prompt includes relevant examples that guide the model's task execution. The model's extensive context window accommodates these prompts and new task-related data, enabling the LLM to recognize patterns and make predictions based on the embedded examples. This ability to identify patterns and draw parallels from its pre-trained knowledge allows the model to effectively perform tasks it was not explicitly trained for, demonstrating the versatility and potential of ICL in diverse applications.

\subsection{Supervised Fine-Tuning}

Supervised fine-tuning (SFT) is a process for adapting a pre-trained LLM to a specific task, enhancing its performance and accuracy \cite{huggingface_sft, wandb_sft}. Following a traditional machine learning workflow,  the process begins with the preparation of a specialized dataset, containing labeled examples pertinent to the task at hand. Starting with a pre-trained model like Llama3 8B, the training process involves adjusting the model's parameters to minimize classification errors, thereby refining its task-specific capabilities. Techniques such as Low-Rank Adaptation (LoRA) are employed to efficiently fine-tune the model by integrating trainable rank decomposition matrices within each layer \cite{lora_arxiv, lora_github}. This reduces the number of trainable parameters, making the process more computationally efficient. Additionally, quantization methods, such as 4-bit quantization, are applied to decrease the model size and inference time while maintaining performance \cite{bnb_github}. Through SFT, LLMs are adapted to meet the demands of specific tasks, unlocking their full potential in real-world applications.

%\subsection{Training Approaches in This Study}
%This study employs two distinct training approaches: ICL and fine-tuning. In the context of this research, training refers to the processes used to adapt the model for the specific task of domain classification.

%For ICL, the training process involves providing the model with a set of examples or prompts during inference. This approach allows the model to adapt its responses based on the given context, leveraging its existing knowledge and contextual understanding without altering its underlying parameters.

%In contrast, fine-tuning involves adjusting the model's parameters using a labeled dataset to improve its performance on the specific task of detecting DGA domains. This process refines the model's capabilities by learning from domain-specific examples, thereby enhancing its accuracy and adaptability to new data.

%Both methods play crucial roles in optimizing the model's ability to classify domains effectively. The selection and implementation of these approaches are fundamental to the study's methodology and results.

\section{Previous Works}
\label{sec:Previos Work}

The study of DGA domain detection and classification has been a dynamic area of research, driven by the ever-evolving tactics of cybercriminals. Initial methodologies were primarily based on lexicographical analysis and heuristic methods. Although these approaches proved effective for straightforward DGAs, they encountered limitations when dealing with more sophisticated algorithms \cite{plohmann2016comprehensive}.

As the field advanced, researchers increasingly adopted machine learning techniques to enhance detection accuracy. These techniques evolved from feature-based approaches, which utilized handcrafted attributes like n-grams, entropy, and character distribution \cite{yu2018character, woodbridge2016predicting}, to the implementation of advanced neural network models. Specifically, Convolutional Neural Networks (CNNs) and Recurrent Neural Networks (RNNs) demonstrated substantial promise in identifying complex patterns in domain names \cite{namgung2022efficient, catania2019analysis, zhou2019cnn}.

In pursuit of improved detection capabilities, hybrid and ensemble approaches have also been explored. An example is the model by Harishkumar and Bhuvaneshwaran \cite{harishkumar2024enhanced}, which integrates n-gram analysis, topic modeling, and attention-based BiLSTM networks. This approach underscores the potential of combining multiple techniques to enhance detection accuracy and robustness.

Continuing advancements in the field address emerging challenges, such as analyzing encrypted traffic. For example, Tapsoba et al. \cite{tapsoba2024analysis} examined DNS over HTTPS (DoH) traffic to identify plaintext features for DGA domain detection, highlighting the potential of leveraging encrypted traffic analysis for cybersecurity applications. Furthermore, the AdamW+ framework \cite{adamwplus} optimizes the AdamW gradient optimizer to improve DGA domain detection.

In a novel approach, AlSabeh et al. \cite{alsabeh2024dga} proposed a framework utilizing P4 programmable switches for DGA detection and classification. This framework leverages the flexibility and processing capabilities of P4 switches to extract unique network heuristics and domain name features through both shallow and deep packet inspection (DPI), with minimal impact on throughput. It employs a two-fold strategy, utilizing a line-rate compact machine learning classifier in the data plane for DGA detection and a comprehensive classifier in the control plane for both detection and classification.

Transformer-based models, recognized for their success in natural language processing tasks, are also being investigated for DGA detection. Yu et al. \cite{dombert} presented Dom-BERT, a model designed to detect malicious domains by constructing a heterogeneous graph and leveraging a pre-trained BERT model. This approach demonstrates significant improvements in F1 score and resilience to class imbalance, but it only uses a small amount of domains. Building upon BERT-based approaches, Mahdaouy et al. introduced DomURLs\_BERT, a specialized pre-trained BERT-based encoder specifically adapted for detecting and classifying suspicious domains and URLs. Their model, which was pre-trained using Masked Language Modeling on a multilingual corpus of URLs, domain names, and DGA datasets, demonstrated superior performance compared to traditional character-based deep learning models and other cybersecurity-focused BERT variants across multiple classification tasks, including phishing, malware, DGA, and DNS tunneling detection \cite{mahdaouy2024domurls_bert}.

The development of sophisticated models has been a hallmark of recent advancements. For instance, Hu et al. \cite{hu2022ci} introduced the CI GRU model, which combines CNNs and gated recurrent units (GRUs) with an attention mechanism. This model significantly enhances the ability to capture temporal dependencies and critical features in DGA domain sequences. Similarly, Tuan et al. \cite{tuan2022detecting} developed the LA\_bin07 model, employing a combination of long short-term memory (LSTM) networks and attention mechanisms to effectively capture both local and global domain name features. This latest model represents the most advanced work in the field so far.

On the other hand, the importance of quality and diverse datasets cannot be overstated in the development and evaluation of DGA detection models \cite{cebere2024down}. Resources such as the UMUDGA dataset \cite{zago2020umudga}, the 360NetLab DGA Dataset \cite{360netlab}, and the Domain Generation Algorithms Repository \cite{dga_repo} provide valuable data for researchers in this field.

Despite these advancements, several challenges persist. Models must quickly adapt to new DGA families without extensive retraining, balance detection accuracy with computational efficiency for real-time applications, and maintain a low false positive rate while achieving a high detection rate.

For their part, the advent of LLMs like GPT-3 has opened new possibilities across various domains, including cybersecurity. Although not extensively explored for DGA detection, LLMs' ability to understand complex linguistic patterns and generalize across domains suggests potential applications in identifying and classifying DGA-generated domains. Recent advancements in LLM architectures and training techniques, such as chain-of-thought prompting \cite{wei2022chain} and reasoning-based approaches \cite{yao2022react}, offer promising avenues for enhancing DGA detection capabilities.

Building upon these foundations, this work investigates how state-of-the-art LLMs can be leveraged to address current challenges in DGA detection. %TP- , potentially offering more accurate, adaptable, and explainable models for cybersecurity applications.

\section{Methodology}
\label{sec:methodology}

\subsection{Dataset Description}
\label{subsec:dataset}
The dataset crafted for this study consists of two primary subsets: one for training and another for testing. Both sets include normal (also known as legit or legitimate) and DGA-generated domains, with the latter distributed across various DGA families \cite{plohmann2016comprehensive}. This assembled dataset provides a comprehensive representation of DGA domains, encompassing various algorithms and strategies employed by different malware families. 

The selection of both normal domains and DGA families to construct the dataset, was conducted following the recommendations of recent studies such as \cite{cebere2024down}, which emphasize the importance of including hard-to-detect families, such as those based on wordlists, to ensure a more robust evaluation of DGA detection methods. The DGA domains were derived from the UMUDGA, DGAarchive, and 360netlab datasets \cite{360netlab, zago2020umudga, plohmann2015dgaarchive}, with additional domains generated following the method proposed in \cite{peck2019charbot}. 

Normal domains were obtained from the Tranco dataset \cite{tranco}. This is a collaborative project that provides a ranking of the most popular websites, aggregating data from various sources to ensure a comprehensive and dynamic list. It is updated regularly, capturing shifts in web popularity and offering a robust reflection of legitimate web activity. %TP- By utilizing this dataset, this study ensures the inclusion of diverse and representative normal domains, which are crucial for accurate evaluation and comparison with DGA domains. %This inclusion supports the study's objective of distinguishing between normal and algorithmically generated domains effectively.

%There were 68 DGA families, of which 14 were separated for generalizability testing. (see section \ref{sec:new_DGA_families}). Therefore, the formed dataset used in the rest of this work includes 54 distinct DGA families, meticulously extracted from the mentioned datasets. Table \ref{tab:domain_distribution} presents the distribution of DGA and normal domains in the training and testing sets, where the DGA were extracted from the 54 families presented in Table \ref{tab:dga_families}, which additionally presents the family category based on its generation scheme: either arithmetic (A) or word-based (W). This classification aids in understanding the underlying mechanisms of domain generation and highlights the diversity of approaches used across different families.

%TP- The dataset used consists of 68 DGA families, along with normal domains. It is important to note that the training set is composed of 54 DGA families, whereas the test set includes these same 54 DGA families, plus 14 additional DGA families used exclusively for testing generalization capabilities on unseen DGA families  (see section \ref{subsec:New_DGA}). These 14 additional families were only used for generalization tests, whereas all other evaluations were conducted using the 54 families that are present in both the training and test sets. For generalization testing, different normal domains were sampled than those used in the other tests.

The dataset used consists of 68 DGA families, along with normal domains. It is important to note that the training set is composed of 54 DGA families, whereas the test set includes these 54 families plus 14 additional ones. These last ones were only used for generalization tests (see Section \ref{subsec:New_DGA}), whereas all other evaluations were conducted using the remaining 54 families. Also for the generalization tests, normal domains different from those used in the other tests were sampled.

Table \ref{tab:domain_distribution} presents the distribution of DGA and normal domains across the training and test sets. The DGA domains were extracted from the 68 families listed in Table \ref{tab:dga_families}, which also categorizes the families based on their generation scheme: arithmetic (A) or word-based (W).  Under the arithmetic scheme, the algorithm usually calculates a sequence of values that have a direct ASCII representation usable for a domain name. On the other hand,  word-based consists of concatenating a sequence of words from one or more wordlists. %TP- This classification helps to better understand the underlying mechanisms of domain generation and highlights the diversity of approaches used across different families.

%TP- The DGA generation scheme followed by the malware families includes  the simple arithmetical (A) and the recent word based (W) schemes.

\begin{table}[ht!]
\centering
\begin{tabular}{lrr}
\toprule
\textbf{Category} & \textbf{Training Set} & \textbf{Testing Set} \\
\midrule
Total DGA domains & 139 million & 15 million \\
Total normal domains & 3 million & 350 thousand \\
Total domains & 142 million & 15.35 million \\
\bottomrule
\end{tabular}
\caption{Distribution of domains in the training and testing sets.}
\label{tab:domain_distribution}
\end{table}

\begin{table*}[ht!]
\centering
\begin{tabular}{llllll}
\toprule
\textbf{Family} & \textbf{Scheme} & \textbf{Family} & \textbf{Scheme} & \textbf{Family} & \textbf{Scheme} \\
\midrule
alureon & A & gozi & W & ramdo & A \\
bamital & A & kraken & A & ramnit & A \\ 
banjori & A & locky & A & ranbyus & A \\
bazarbackdoor* & A & manuelita & W & rovnix & W \\
bedep & A & matsnu & W & sharkbot* & A \\ 
bigviktor* & W & monerominer & A & shiotob & A \\
bumblebee* & A & murofet & A & simda & A \\
ccleaner* & A & murofetweekly & A & sisron & A \\
charbot & W & mydoom & A & sphinx & A \\
chinad & A & necurs & A & suppobox & W \\
conficker & A & new\_goz* & A & symmi & A \\
corebot & A & ngioweb* & W & tempedreve & A \\
cryptolocker & A & nymaim & W & tinba & A \\
deception & W & oderoor & A & tinynuke & A \\
dircrypt & A & padcrypt & A & tufik* & A \\
dmsniff* & A & pitou & A & vawtrak & A \\
dnschanger & A & pizd* & W & verblecon* & A \\ 
dyre & A & proslikefan & A & vidro & A \\
emotet & A & pushdo & A & virut & A \\
enviserv* & A & pykspa & A & xshellghost* & A \\
fobber & A & qadars & A & zeus-newgoz & A \\
gameover & A & qakbot & A & zloader & A \\
goz* & A & qsnatch & A & & \\
\bottomrule
\end{tabular}
\caption{DGA families and their schemes used in the dataset for training and testing. (A) stands for Arithmetic generation scheme, (W) for Word-based. (*) indicates the 14 additional families in the test set.}
\label{tab:dga_families}
\end{table*}

%TP- Normal domains were obtained from the Tranco dataset \cite{tranco}. It is a collaborative project that provides a ranking of the most popular websites, aggregating data from various sources to ensure a comprehensive and dynamic list. The dataset is updated regularly, capturing shifts in web popularity and offering a robust reflection of legitimate web activity. By utilizing the Tranco dataset, this study ensures the inclusion of diverse and representative normal domains, which are crucial for accurate evaluation and comparison with DGA domains. This inclusion supports the study's objective of distinguishing between normal and algorithmically generated domains effectively.

It is worth noting that the training and testing datasets described above were not used in their totality, but rather different samples were taken from them to evaluate distinct approaches, as described below. 

\subsection{Training Process}
\label{subsec:training-process}
%% Training definition
In the context of this research, the term \textit{training} refers to the processes used to adapt the model for the specific task of domain classification. In particular, this study employs two distinct training approaches: ICL and SFT, as described in Section ~\ref{sec:background}.

%% methodologies
%TP- The Llama3 8B model was trained using different methodologies with data extracted from the training dataset, as illustrated in Figure \ref{fig:distribution_data_train}. 

%To achieve optimal performance and resource efficiency, the fine-tuning process employed several advanced techniques and tools:

%\textit{SFTTrainer}: This tool facilitated the training process by applying LoRA and the quantization configuration to the prepared datasets \cite{huggingface_sft, wandb_sft}

%The \textit{Data\-Collator\-For\-Completion\-Only\-LM} ensured that each example was presented in a suitable format for the model's learning process \cite{huggingface_data_collator}.

\vspace{0.5cm}
%TP- As depicted in Figure \ref{fig:distribution_data_train}, the Llama3 8B model was trained using different methodologies with data sampled from the training dataset.  In all cases, the DGA instances were sampled from 54 available families. However, this study employs specific quantities of domain data, depending on the training approach. For the SFT method, a substantial dataset of two million domains was utilized, evenly split between one million DGA domains and one million normal domains. Analogously, two separate dataset were used for the ICL-based models: one using a sample of 500 domains (250 DGA and 250 normal), and another utilizing a larger set of 2000 domains (1000 DGA and 1000 normal). Notice that the number of domains used in ICL approaches were constrained by the context window. 

\begin{figure}[ht]
    \centering
    \includegraphics[width=\textwidth]{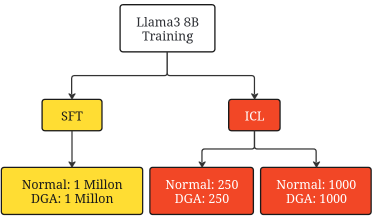}
    \caption{Distribution of domains for the different training methods.}
    \label{fig:distribution_data_train}
\end{figure}

As depicted in Figure \ref{fig:distribution_data_train}, the Llama3 8B model was trained using different methodologies with data sampled from the training dataset.  However, this study employs specific quantities of domain data, depending on the training approach. For the SFT method, a substantial dataset of two million domains was utilized, evenly split between one million DGA domains and one million normal domains. Analogously, two separate datasets were used for the ICL-based models: one using a sample of 500 domains (250 DGA and 250 normal), and another utilizing a larger set of 2000 domains (1000 DGA and 1000 normal). Notice that the number of domains used in ICL approaches was constrained by the context window. In all cases, the DGA instances were sampled from the 54 available families, so for example, approximately 18 DGA domains per family were sampled to obtain the 1000 DGA domains for the second dataset used in the ICL training approach. 

%TP- This variation in batch size for ICL allows for a comparative analysis of model performance under different data volume conditions (see Figure~\ref{fig:distribution_data_train}).

A consistent data format was used for SFT, ensuring that the model received the data properly to learn the relationships between domains and their corresponding labels. Each example was structured as in Figure \ref{fig:example_format}.

\begin{figure}[ht]
    \centering
    \begin{tcolorbox}[colback=gray!10, colframe=gray!50, title=Example Format]
    \begin{lstlisting}
    #domain: {domain}
    #label: {label}
    \end{lstlisting}
    \end{tcolorbox}
    \caption{Example format for training data (for SFT), including domain and label.}
    \label{fig:example_format}
\end{figure}

%In contrast, the ICL training approach used with the Llama3 8B model employed significantly smaller datasets. Two separate ICL models were trained: the first using a sample of 500 domains (250 DGA and 250 normal), and the second utilizing a larger set of 2000 domains (1000 DGA and 1000 normal). This variation in dataset sizes for ICL allows for a comparative analysis of model performance under different data volume conditions.

In contrast, when ICL was applied, the model was iteratively presented with a prompt (Figure \ref{Prompt}) outlining its classification task, followed by the 500 or 2000 labeled domain names, as well as a domain name extracted from the testing set. This approach enabled the model to classify new domain names based on the enhanced knowledge gained from the prompt, demonstrating ICL's effectiveness in addressing domain classification challenges. 

%TP- Two separate ICL-based models were trained: one using a sample of 500 domains (250 DGA and 250 normal), and another utilizing a larger set of 2000 domains (1000 DGA and 1000 normal). Note that DGA instances were sampled from all available families. This variation in batch size for ICL allows for a comparative analysis of model performance under different data volume conditions (see Figure~\ref{fig:distribution_data_train}).

\begin{figure*}[ht!]
\centering
\begin{tcolorbox}[colback=yellow!5!white,colframe=yellow!75!black, title=Prompt used for domain classification, width=\textwidth] % width=\textwidth
\begin{verbatim}
You are a domain name classification system. Your task is
to classify domain names as either 'dga' (Domain Generation
Algorithm) or 'normal'. DGA domains are automatically
generated by malware, while normal domains are not. I will
provide you with labeled training data containing domain
names and their classifications. After the training phase,
you will classify a new domain and respond with either
'dga' or 'normal'.
as.com
domain: as.com, result: normal
...
...
...
xcfdreyjs.com
domain: xcfdreyjs.com, result: dga
Now you classify this domain: google.com, only answer
dga or normal. Do not provide any additional information
or explanation.
\end{verbatim}
\end{tcolorbox}
\caption{Prompt used on Llama3 8B for classification of domain names in ICL.}
\label{Prompt}
\end{figure*}
%The experiment evaluated the model's ability to classify domain names using batches containing 500  (250 DGA and 250 normal) and 2000 (1000 DGA and 1000 normal) instances, labeled as legitimate or malicious. These dataset batches acted as prompts, facilitating the identification of domain classification patterns from the provided examples. The context window size played a crucial role in ensuring the model processed information efficiently and delivered accurate classification results.

%%% memory Details
For applying ICL, the setup involved configuring the model locally using the \texttt{ollama} package \cite{ollama}, which provides an optimized interface for working with language models, allowing specific adjustments and configurations to suit the experiment's requirements. The model features a quantization type of Q4\_0 and quantization version 2 for dealing with hardware limitations. On the other hand, during the SFT process, the Low-Rank Adaptation (LoRA) technique ~\cite{lora_arxiv, lora_github, dettmers2024qlora} was used during training to deal with hardware limitations. This technique was applied to the key, value, and query projection modules (\texttt{k\_proj}, \texttt{v\_proj}, \texttt{q\_proj}) of the decoder layers. LoRA allows for the modification of specific model components, thereby reducing the need to adjust all parameters and decreasing training complexity \cite{lora_arxiv, lora_github}. In addition, the model underwent 4-bit quantization during training. This approach significantly decreases memory usage while preserving model performance \cite{bnb_github}.

\subsection{Evaluation Process}
\label{subsec:performance-evaluation}

\begin{figure}[ht]
    \centering
    \includegraphics[width=1\columnwidth]{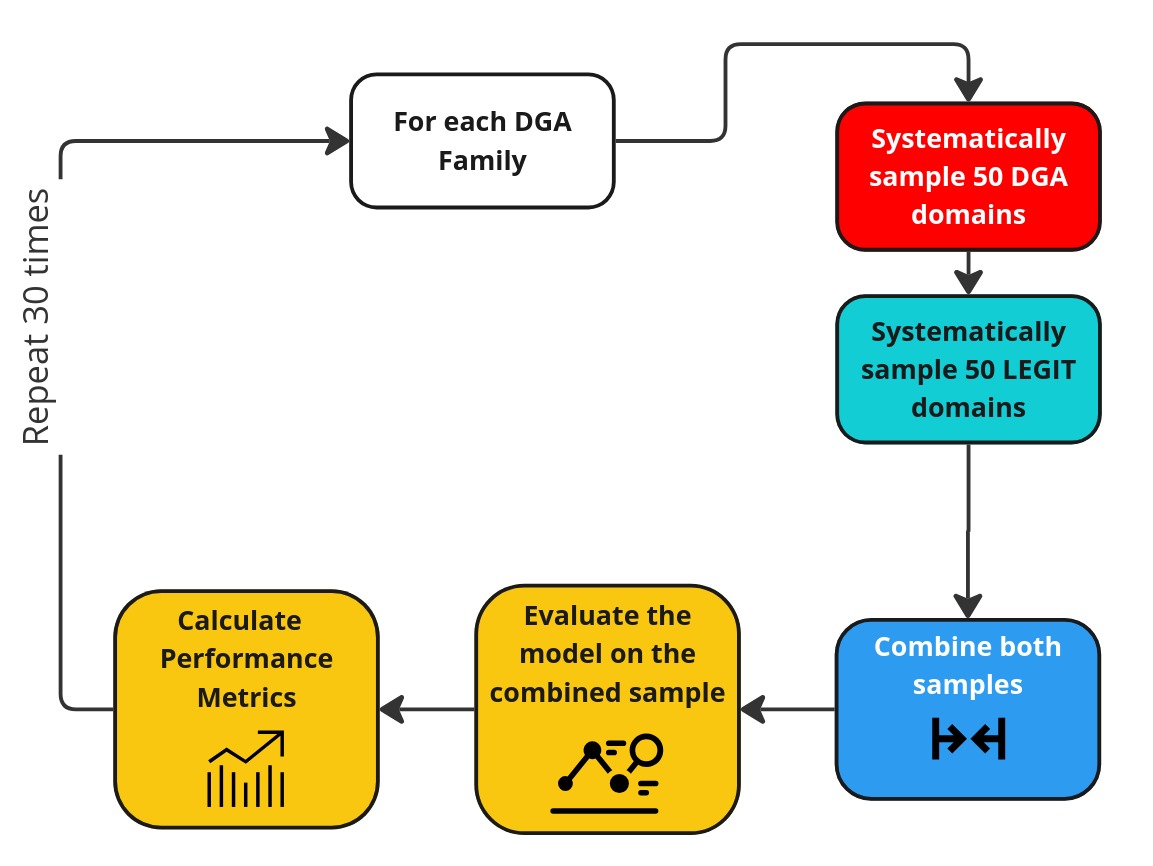}
    \caption{Model evaluation diagram.}
    \label{fig:diagram_test}
\end{figure}

Regardless of their training approach, a consistent methodology was applied across all the models.  The flowchart in Figure \ref{fig:diagram_test} illustrates the evaluation procedure step by step, highlighting the systematic approach adopted in the study.

To ensure the reliability and robustness of the evaluation results, thirty independent runs were conducted \textit{for each  DGA family}. On each run, a systematic sample~\cite{cochran1977sampling} of 100 example domains (50 DGA and 50 legit) was provided to the model for classification into DGA-generated or legitimate. The model predictions, query times, and other relevant performance metrics were recorded throughout these runs to establish its effectiveness in detecting malicious domains generated by DGAs. Finally, the performance of the models was assessed using several key metrics, each of which provides insights into different aspects of model efficacy. The metrics utilized in this evaluation include Accuracy (Accu), Precision (Pre), Recall (Re), F1 Score (F1), False Positive Rate (FPR), and Processing Time (Proc. Time) \cite{hossin2015review, powers2020evaluation}.

%In summary, this comprehensive procedure involved organizing the DGA domain families into separate files, sampling domains from each family, combining these samples with legitimate domains, and repeatedly testing the model is ability to identify DGA domains across multiple runs. 
%This approach ensured a robust evaluation of the model's performance in detecting malicious domains generated by various DGA families. 

%TP- A more detailed explanation of the systematic sampling methodology is provided to clarify the specific approach used in this process.

%TP- The systematic sampling strategy applied, as visually described in Figure~\ref{fig:systematic-sampling}, was as follows: samples were selected from the DGA and the legit domains in the test set at thirty regular intervals, each of length fifty. Similarly, fifty legitimate domains were also sampled from the test set at each of the thirty intervals, resulting in a final sample of one hundred domains per interval. Notice that, for each family, the same 30 legit sample domains were combined with the corresponding 30 DGA domains.

\begin{figure}[h]
    \centering
    \includegraphics[width=1\columnwidth]{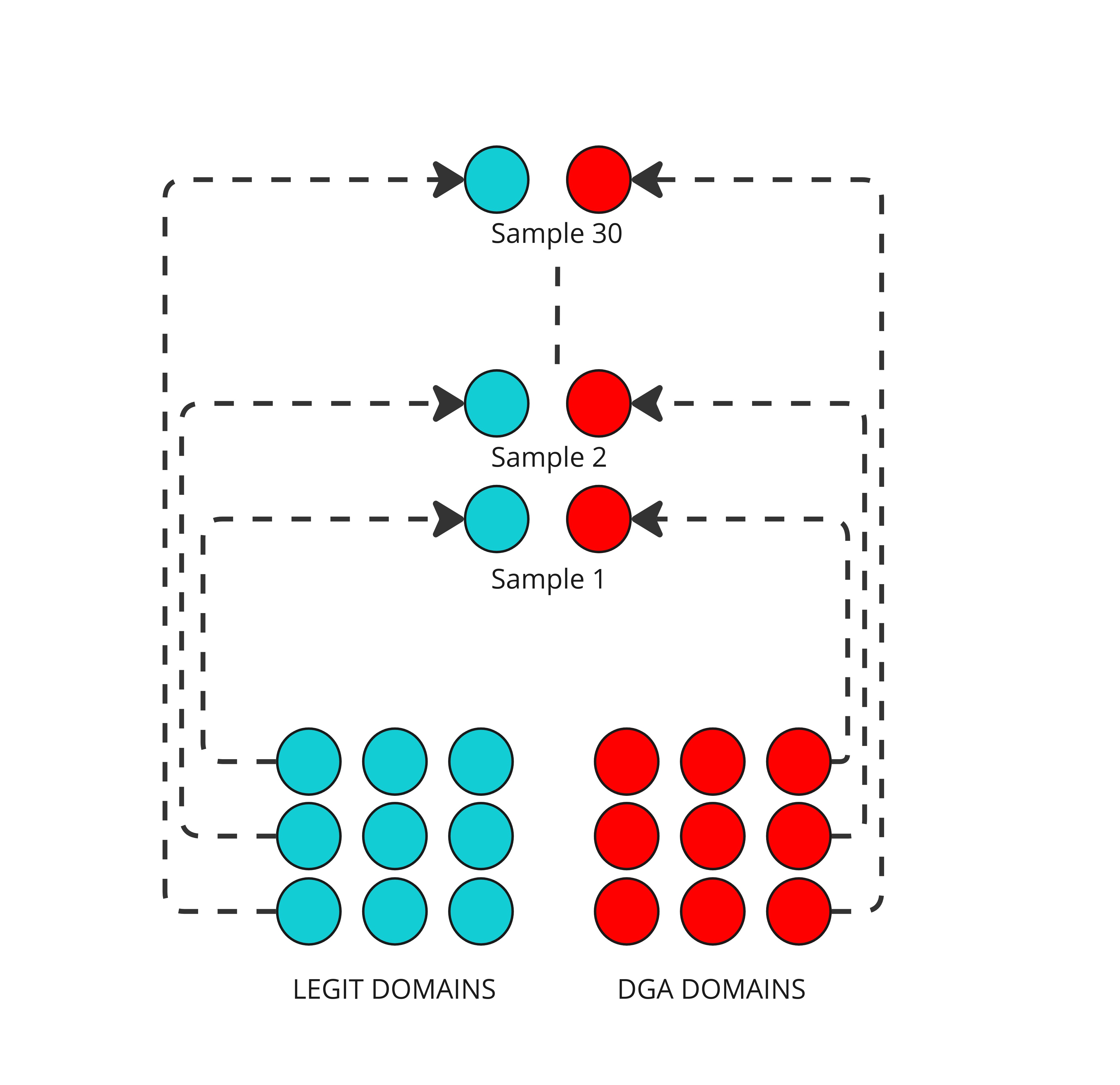}
    \caption{Systematic sampling: 30 samples of 50 legit and 50 DGA domains. Each circle represents 50 different domains, and all circles are disjoint.}
    \label{fig:systematic-sampling}
\end{figure}

The systematic sampling strategy applied, as visually described in Figure~\ref{fig:systematic-sampling}, was as follows: samples were selected from the DGA and the legitimate domains in the test set at thirty regular intervals, each one of length fifty. In the end, there were one hundred domains per interval: fifty DGA and fifty legit. Notice that, for each family, the same thirty intervals of legitimate domains were combined with the corresponding thirty of DGA.%TP- the same 30 legit sample domains were combined with the corresponding 30 DGA domains.

%In particular, the systematic sampling strategy~\cite{cochran1977sampling} applied was the following,  after shuffling, samples were selected from the DGA domains of the test set at 30 regular intervals of length 50. Analogously, 50 legit domains were also sampled from the test set for each of the 30 intervals. Then, for every family, the same 1500 legit domains were combined with the corresponding 1500 DGA domains, divided into 30 groups of 100 domains (50 DGA and 50 legitimate). 

\section{Experimental Results}
\label{sec:Experiment}
This section outlines the results of the effectiveness of LLMs in detecting DGA across a series of experiments. First, we explored the \textbf{optimal training strategy} for LLMs in DGA detection, comparing ICL and SFT to determine which approach performs better. Next, we assessed their ability to detect malicious domains when tested on new examples of DGA families present in the training data, examining how well the model distinguishes between DGA and legitimate domains. Then we extend the problem by evaluating the generalization capabilities of the SFT model by testing its performance on previously unseen DGA families to understand how well it handles new domain generation algorithms. Finally, we compared the SFT-trained LLM to state-of-the-art models, such as LA\_Bin07, a deep learning model using attention mechanisms. 
% Evaluate each approach from section LLM approaches 
% Use two models: binary   (dga/normal) 
To ensure reproducibility, all the source code, datasets, and instructions used in this study are available in a public GitHub repository \cite{reypapin2024dga}.

%\subsection{ICL}

\subsection{Experiment I: Evaluation of Training Approaches } \label{sec:comparison-of-training-approaches}

%TP- The results from this experiment used domains from the same 54 families, which were also present in the train set.

The results of this experiment used domains from the 54 families present in the training set.

%TP- Table \ref{tab:comparison_metrics_sample_sizes} presents the results of the ICL training process as described in section~\ref{subsec:training-process}. The table illustrates how training with different dataset sizes impacts the model's ability to distinguish between legitimate and malicious domains.

Table \ref{tab:comparison_metrics_sample_sizes} presents the overall average of the metrics obtained, for all runs and the 54 families, after applying the ICL training process described in Section~\ref{subsec:training-process}. The table illustrates how training with different dataset sizes affects the model's ability to distinguish between legitimate and malicious domains.

%\begin{table}[ht!]
%\centering
%\begin{tabular}{lrr}
%\toprule
%\textbf{Metric} & \multicolumn{2}{c}{\textbf{Llama3 8B ICL}} \\  % Agregamos multicolumn para centrar el texto
%\cmidrule(lr){2-3}  % Dibuja una línea bajo la celda que abarca las dos columnas
% & \textbf{2000 Samples} & \textbf{500 Samples} \\
%\midrule
%Accuracy & 0.84 & 0.55 \\
%Precision & 0.87 & 0.52 \\
%Recall & 0.78 & 0.99 \\
%F1 Score & 0.81 & 0.69 \\
%FPR & 0.09 & 0.88 \\ 
%Proc. Time & 1.47 & 0.95 \\
%\bottomrule
%\end{tabular}
%\caption{Performance comparison of Llama3\_8B model with different sample sizes.}
%\label{tab_met}
%\end{table}

\begin{table}[ht!]
\centering
\fontsize{8}{9}\selectfont
\begin{tabular}{ccccccc}
\toprule
\textbf{Sample Size} & \textbf{Accu} & \textbf{Pre} & \textbf{Re} & \textbf{F1} & \textbf{FPR} &\textbf{ Proc. Time (s)} \\
\midrule
2000 & 0.84 & 0.87 & 0.78 & 0.81 & 0.1 & 1.47 \\
500 & 0.55 & 0.52 & 0.99 & 0.69 & 0.88 & 0.95 \\
\bottomrule
\end{tabular}
\caption{Performance comparison of Llama3 8B model with different sample sizes for training, using the ICL approach.}
\label{tab:comparison_metrics_sample_sizes}
\end{table}

The model using 500 domains shows poor performance, with a recall value of 0.99 but a precision of 0.52. The model is not capable of distinguishing between DGA and legitimate domains. In addition, the FPR is extremely high.  

On the other hand, the model trained with 2000 domains outperforms the previous one in all evaluation metrics except processing time and recall. Notably is the increment in the precision (0.87) and the diminution of the FPR (0.1). Performance analysis of this model is presented in Table \ref{table:metric-llama3-L2K}, averaging by family, which highlights the strengths and weaknesses of the model for each DGA family. %TP- This table highlights the strengths and weaknesses of the model for each DGA family, offering insights into its effectiveness in differentiating between legitimate and malicious domains within each family.

\begin{table*}[ht!]
\centering
\small
\begin{tabular}{lc c c|lc c c}
\toprule
\multicolumn{1}{c}{\textbf{Family}} & \textbf{Pre} & \textbf{Re} & \textbf{F1} & \multicolumn{1}{c}{\textbf{Family}} & \textbf{Pre} & \textbf{Re} & \textbf{F1} \\ \midrule
alureon & 0.91 & 0.95 & 0.93 & oderoor & 0.90 & 0.79 & 0.84 \\
bamital & 0.90 & 0.92 & 0.91 & padcrypt & 0.91 & 0.95 & 0.93 \\
banjori & 0.90 & 0.87 & 0.88 & pitou & 0.87 & 0.68 & 0.76 \\
bedep & 0.91 & 1.00 & 0.95 & proslikefan & 0.86 & 0.60 & 0.70 \\
charbot & 0.80 & 0.36 & 0.49 & pushdo & 0.90 & 0.88 & 0.89 \\
chinad & 0.91 & 0.99 & 0.95 & pykspa & 0.87 & 0.66 & 0.75 \\
conficker & 0.86 & 0.61 & 0.72 & qadars & 0.91 & 0.93 & 0.92 \\
corebot & 0.91 & 1.00 & 0.95 & qakbot & 0.91 & 0.93 & 0.92 \\
cryptolocker & 0.91 & 0.95 & 0.93 & qsnatch & 0.71 & 0.26 & 0.38 \\
deception & 0.84 & 0.49 & 0.62 & ramdo & 0.91 & 0.93 & 0.92 \\
dircrypt & 0.91 & 0.97 & 0.94 & ramnit & 0.91 & 0.96 & 0.93 \\
dnschanger & 0.92 & 0.95 & 0.93 & ranbyus & 0.91 & 1.00 & 0.95 \\
dyre & 0.90 & 0.90 & 0.90 & rovnix & 0.87 & 0.64 & 0.73 \\
emotet & 0.91 & 0.96 & 0.93 & shiotob & 0.91 & 0.96 & 0.93 \\
fobber & 0.91 & 1.00 & 0.95 & simda & 0.85 & 0.52 & 0.65 \\
gameover & 0.91 & 1.00 & 0.95 & sisron & 0.89 & 0.83 & 0.86 \\
gozi & 0.89 & 0.80 & 0.84 & sphinx & 0.91 & 1.00 & 0.95 \\
kraken & 0.90 & 0.72 & 0.80 & suppobox & 0.62 & 0.15 & 0.24 \\
locky & 0.91 & 0.87 & 0.89 & symmi & 0.91 & 1.00 & 0.95 \\
manuelita & 0.72 & 0.25 & 0.37 & tempedreve & 0.89 & 0.82 & 0.85 \\
matsnu & 0.64 & 0.15 & 0.24 & tinba & 0.90 & 0.95 & 0.93 \\
monerominer & 0.87 & 0.63 & 0.73 & tinynuke & 0.90 & 0.86 & 0.88 \\
murofet & 0.91 & 0.98 & 0.94 & vawtrak & 0.89 & 0.79 & 0.84 \\
murofetweekly & 0.92 & 1.00 & 0.96 & vidro & 0.90 & 0.85 & 0.87 \\
mydoom & 0.89 & 0.75 & 0.81 & virut & 0.79 & 0.37 & 0.50 \\
necurs & 0.90 & 0.90 & 0.90 & zeus-newgoz & 0.92 & 1.00 & 0.96 \\
nymaim & 0.75 & 0.28 & 0.40 & zloader & 0.91 & 0.97 & 0.94 \\
\bottomrule
\end{tabular}
\caption{Metrics obtained with the Llama3 8B model trained using the ICL approach (with 2000 samples).}
\label{table:metric-llama3-L2K}
\end{table*}

Table \ref{table:metric-llama3-L2K} shows that the model exhibits high precision. Furthermore, for many DGA families, the model achieves scores greater than or equal to 0.9 in all metrics. Notably, families such as \texttt{murofetweekly}, \texttt{dnschanger}, and \texttt{zeus-newgoz} achieve a precision of approximately 0.92. However, variability is observed, with families like \texttt{manuelita}, \texttt{matsnu}, \texttt{qsnatch}, and \texttt{suppobox} showing lower performance, with a precision around 0.65. These less accurately detected families predominantly consist of word-based domains, which are generally more challenging for detectors.

%\todo[inline]{review these paragraphs} Resolved

%TP- Overall, the results demonstrate that the ICL-trained Llama3 8B model performs acceptably on this domain classification task while highlighting areas for improvement for several DGA families. With only 18 samples per family size, the LLM can get more complete information about the patterns that distinguish legitimate and malicious domains, thus improving its generalization ability. 

%RL- Overall, the results demonstrate that the Llama3 8B model trained using the ICL approach with 2000 domains, performs acceptably on the domain classification task while highlighting areas for improvement for several DGA families. That is, the LLM can get information about patterns that distinguish legitimate and malicious domains, and this can be achieved with approximately 18 example domains per family, to form the set of 1000 DGA domains. This set is then combined with another 1000 legitimate domains, allowing the model to learn and differentiate between both types of domains.

Overall, the results demonstrate that the Llama3 8B model, trained using the ICL approach with 2000 domains, performs acceptably on the domain classification task, while also highlighting areas for improvement in several DGA families. Specifically, the LLM can identify patterns that distinguish legitimate domains from malicious ones, and this can be achieved with approximately 18 example domains per DGA family. %TP- , forming a set of 1000 DGA domains. This set is then combined with another 1000 legitimate domains, allowing the model to learn and differentiate between the two types of domains.

%Conversely, the model trained with 500 domains displays lower performance across all metrics, likely due to the limited information in the smaller sample, which constrains the model's ability to capture diverse patterns in DGA families. 

%The model trained with 2000 domains shows superior accuracy, precision, recall, and F1 scores compared to the model trained with 500 domains, underscoring the advantages of a larger training dataset.

These findings confirm that a larger sample size enhances the performance of LLMs in domain classification tasks using ICL, emphasizing the importance of providing sufficient context for optimal results.

%\vspace{1.0cm}

%\subsection{SFT}

%TP- On the other hand, the SFT results of the Llama3 8B model are shown in Table~\ref{table:sumary-metric}. The SFT Llama3 8B model demonstrated high accuracy (94\%), precision (93\%), recall (92\%), and F1 score (92\%), excelling in correctly identifying both positive and negative examples. It also has a low false positive rate (FPR) of 0.04. This low false positive rate is crucial in applications where minimizing false positives is essential, making it a valuable model despite its processing time.

On the other hand, the overall average over all runs and families, of the metrics obtained after applying the SFT training approach, are shown in Table~\ref{table:sumary-metric}. The SFT Llama3 8B model demonstrated high accuracy (94\%), precision (93\%), recall (92\%), and F1 score (92\%), excelling in correctly identifying both positive and negative examples. It also has a low FPR of 0.04. The latter is crucial in applications where minimizing false positives is essential, making it a valuable model despite its processing time.

\begin{table}[ht!]
\centering
\fontsize{8}{9}\selectfont
\begin{tabular}{lcccccc}
\toprule
\textbf{Model} & \textbf{Accu} & \textbf{Pre} & \textbf{Re} & \textbf{F1} & \textbf{FPR} &\textbf{ Proc. Time (s)} \\
\midrule
SFT Llama3 8B & 0.94 & 0.93 & 0.92 & 0.92 & 0.04 & 3.50 \\
\bottomrule
\end{tabular}
\caption{Llama3 8B model performance, trained using the SFT approach.}
\label{table:sumary-metric}
\end{table}

Table \ref{table_Llama3_Pre,F1,Re} presents the evaluation metrics for the SFT Llama3 8B model across various DGA families. The results demonstrate the model's effectiveness in classifying DGA families, with different levels of precision and recall observed across the families. This detailed analysis is crucial for optimizing model performance and effectively addressing the unique challenges associated with each DGA family.

\begin{table*}[ht!]
\centering
\small
\begin{tabular}{lccc|lccc}
\toprule
\multicolumn{1}{c}{\textbf{Family}} & \textbf{Pre} & \textbf{Re} & \textbf{F1} & \multicolumn{1}{c}{\textbf{Family}} & \textbf{Pre} & \textbf{Re} & \textbf{F1} \\ \midrule
alureon        & 0.96 & 1.00 & 0.98 & oderoor        & 0.95 & 1.00 & 0.97 \\
bamital        & 0.96 & 1.00 & 0.98 & padcrypt       & 0.96 & 1.00 & 0.98 \\
banjori        & 0.96 & 0.98 & 0.97 & pitou          & 0.95 & 0.90 & 0.92 \\
bedep          & 0.95 & 1.00 & 0.98 & proslikefan    & 0.96 & 0.96 & 0.96 \\
charbot        & 0.95 & 0.81 & 0.87 & pushdo         & 0.95 & 0.95 & 0.95 \\
chinad         & 0.96 & 1.00 & 0.98 & pykspa         & 0.96 & 0.95 & 0.96 \\
conficker      & 0.95 & 0.84 & 0.89 & qadars         & 0.96 & 0.99 & 0.98 \\
corebot        & 0.96 & 1.00 & 0.98 & qakbot         & 0.96 & 1.00 & 0.98 \\
cryptolocker   & 0.95 & 1.00 & 0.98 & qsnatch        & 0.88 & 0.40 & 0.54 \\
deception      & 0.95 & 0.87 & 0.91 & ramdo          & 0.96 & 1.00 & 0.98 \\
dircrypt       & 0.96 & 0.99 & 0.98 & ramnit         & 0.96 & 0.99 & 0.97 \\
dnschanger     & 0.95 & 0.99 & 0.97 & ranbyus        & 0.96 & 1.00 & 0.98 \\
dyre           & 0.95 & 1.00 & 0.98 & rovnix         & 0.96 & 0.88 & 0.92 \\
emotet         & 0.96 & 1.00 & 0.98 & shiotob        & 0.96 & 0.97 & 0.96 \\
fobber         & 0.96 & 1.00 & 0.98 & simda          & 0.96 & 1.00 & 0.98 \\
gameover       & 0.95 & 1.00 & 0.98 & sisron         & 0.96 & 1.00 & 0.98 \\
gozi           & 0.95 & 0.97 & 0.96 & sphinx         & 0.96 & 1.00 & 0.98 \\
kraken         & 0.95 & 1.00 & 0.98 & suppobox       & 0.95 & 0.92 & 0.94 \\
locky          & 0.95 & 0.99 & 0.97 & symmi          & 0.95 & 1.00 & 0.98 \\
manuelita      & 0.87 & 0.29 & 0.43 & tempedreve     & 0.96 & 1.00 & 0.98 \\
matsnu         & 0.95 & 0.79 & 0.86 & tinba          & 0.95 & 1.00 & 0.98 \\
monerominer    & 0.95 & 0.97 & 0.96 & tinynuke       & 0.01 & 0.00 & 0.00 \\
murofet        & 0.95 & 1.00 & 0.98 & vawtrak        & 0.96 & 0.91 & 0.93 \\
murofetweekly  & 0.96 & 1.00 & 0.98 & vidro          & 0.96 & 0.98 & 0.97 \\
mydoom         & 0.95 & 1.00 & 0.97 & virut          & 0.94 & 0.66 & 0.77 \\
necurs         & 0.96 & 0.99 & 0.97 & zeus-newgoz    & 0.96 & 1.00 & 0.98 \\
nymaim         & 0.95 & 0.90 & 0.93 & zloader        & 0.96 & 1.00 & 0.98 \\ 
\bottomrule
\end{tabular}
\caption{Metrics obtained with the Llama3 8B model, trained using the SFT approach.}
\label{table_Llama3_Pre,F1,Re}
\end{table*}

%The SFT results highlight the potential of advanced language models in cybersecurity applications, specifically in identifying and mitigating threats posed by DGAs.

%This section provides a detailed comparison of the performance between the Llama3 8B model fine-tuned with domain-specific data and the model employing ICL. 

%\subsubsection{Performance Metrics and Results}

%\begin{table}[ht!]
%\centering
%\begin{tabular}{lcc}
%\toprule
%\textbf{Metric} & \textbf{ICL 2000 Sample} & \textbf{Fine Tuning} \\
%\midrule
%Accuracy & 0.84 & 0.94 \\
%Precision & 0.87 & 0.93 \\
%Recall & 0.78 & 0.92 \\
%F1 Score & 0.81 & 0.92 \\
%FPR & 0.10 & 0.04 \\ 
%Proc. Time (s) & 1.47 & 3.50 \\
%\bottomrule
%\end{tabular}
%\caption{Performance comparison of Llama3 8B model ICL 2000 sample vs Llama3 8B Fine Tuning.}
%\label{tab_met_ICLvsFT}
%\end{table}

\begin{figure}[ht!]
    \centering
    \includegraphics[width=\columnwidth]{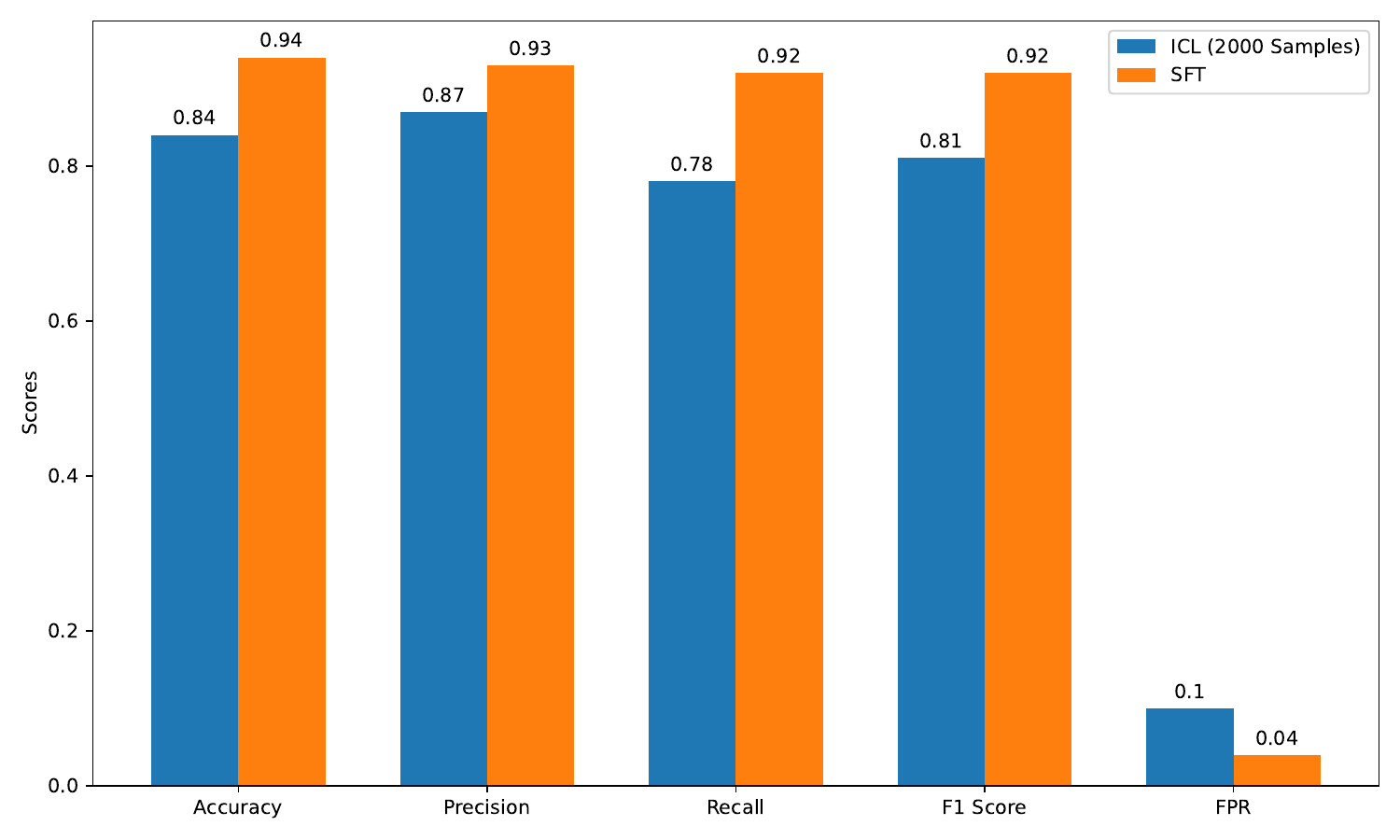} 
    %TP- \caption{Performance comparison of Llama3 8B model ICL 2000 sample vs Llama3 8B Fine Tuning}
    \caption{Performance comparison of Llama3 8B model, using the ICL (with 2000 samples) and the SFT training approaches.}
    \label{fig:comparison_metrics_Llama3FT-ICL}
\end{figure}

When comparing the SFT with ICL models, the experiments showed that the SFT Llama3 8B model significantly outperforms the ICL-based approach across multiple evaluation metrics, as evidenced in Figure \ref{fig:comparison_metrics_Llama3FT-ICL}. %SFT model outperformed the ICL-based approach in all evaluation metrics: detection accuracy (94\% vs 84\%), precision (93\% vs 87\%), recall (92\% vs 78\%), F1 score (92\% vs 81\%), and false positive rate (4\% vs 10\%). These  results indicate that the SFT model is more effective, reliable, and trustworthy for detecting DGA domains.

%TP- Figure \ref{fig:comparison_metrics_Llama3FT-ICL} presents the comparative performance of the Llama3 8B model using ICL versus SFT based on the F1\_score metric for all the DGA families. The mean F1\_score values for each family reveal that the SFT model consistently outperforms the ICL model, achieving near perfect F1\_score in most cases. The ICL model demonstrates greater variability, with significant drops in F1\_score for certain families, such as \textit{charbot, conficker, deception, matsnu, manuelita, nymaim, suppobox, virut}, and \textit{simda}. In contrast, the SFT model maintains a high F1\_score across nearly all families, with only minor fluctuations. These results underscore the robustness and reliability of the SFT model in accurately detecting DGA domains across a diverse set of families, making it a more effective approach for DGA detection.

\begin{figure*}[ht]
    \centering
    \includegraphics[width=\textwidth]{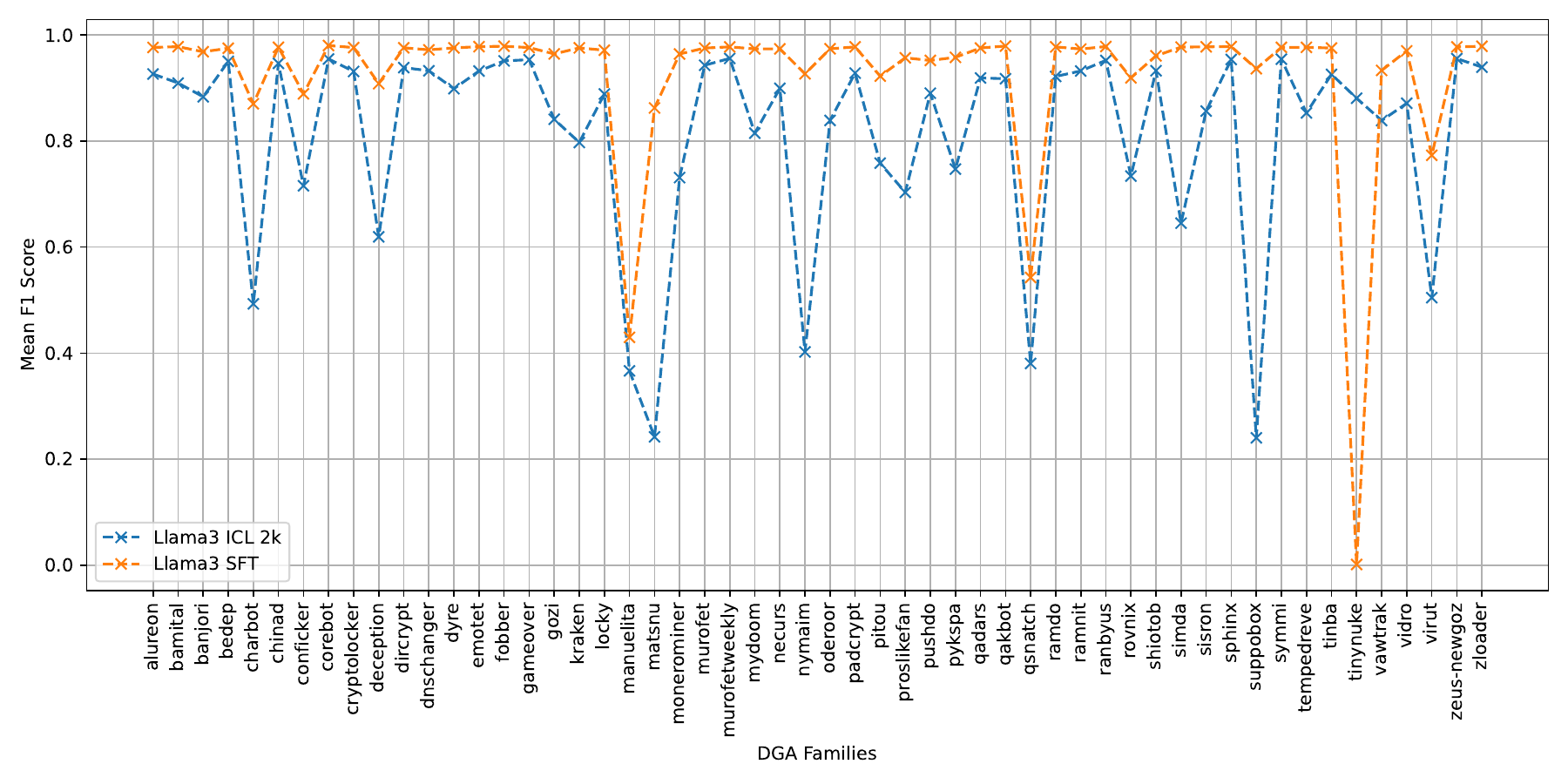}
    %TP- \caption{Comparison of F1\_score between ICL 2000 sample and SFT models}
    \caption{F1 score comparison of Llama3 8B model, using the ICL (with 2000 samples) and the SFT training approaches.}
    
    \label{fig:comparison_metrics_Llama3FT-ICL-family}
\end{figure*}

Figure \ref{fig:comparison_metrics_Llama3FT-ICL-family} presents the mean F1 score, obtained with the Llama3 8B model using ICL (with 2000 samples) and SFT training approaches, for the 54 DGA families. It reveals that the SFT model consistently outperforms the ICL model, achieving a near-perfect F1 score in most cases. The ICL model demonstrates greater variability, with significant drops in F1 score for certain families, such as \texttt{charbot, deception, manuelita, matsnu, nymaim, qsnatch, simda, suppobox}, and \texttt{virut}, all with a mean F1 score lower than or equal to 0.65. In contrast, the SFT model maintains a high F1 score across nearly all families, with only minor fluctuations. These results underscore the robustness and reliability of the SFT model in detecting DGA domains across a diverse set of families, making it a more effective approach for DGA detection.

% \begin{figure*}[ht]
%     \centering
%     \includegraphics[width=\textwidth]{Compartion_Llama3_ICLvsSFT.png}
%     %TP- \caption{Comparison of F1\_score between ICL 2000 sample and SFT models}
%     \caption{F1 score comparison of Llama3 8B model, using the ICL (with 2000 samples) and the SFT training approaches.}
    
%     \label{fig:comparison_metrics_Llama3FT-ICL-family}
% \end{figure*}

Based on these results, we can conclude that the SFT Llama3 8B model outperforms in detecting DGA domains. Therefore, for the remainder of this study, only this model will be used to ensure optimal metrics and efficiency in DGA detection.

\subsection{Experiment II: Evaluation of Generalization to New Families}
\label{subsec:New_DGA}

The generalization capability of a machine learning model to new, unseen data is crucial for its practical application in real-world scenarios. This is particularly significant in cybersecurity, where new threats and variations of existing threats constantly emerge. 

To assess the generalization capability of the SFT Llama3 8B model, a comprehensive evaluation was conducted. The model was tested on a set of 14 previously unseen DGA families, representing a diverse range of DGAs. These new DGA families are found only in the test set, as described in Section \ref{subsec:dataset}. Also, new normal domains were used for this test. This evaluation aims to provide a thorough understanding of the model's strengths and limitations when faced with novel data. These 14 families were evaluated following the procedure outlined in Section \ref{subsec:performance-evaluation}.

%The evaluation focused on a diverse set of fourteen DGA families that were not used during the model is training phase. These families were categorized into two main types: those generated by word lists (W) and those generated arithmetically (A) through various computational methods such as algorithms, permutations, or hashing techniques. This classification allows for a more nuanced analysis of the model's performance across different DGA generation strategies.

%Table \ref{table:generalization} presents the fourteen new DGA families along with their classification according to their generation algorithm. 

% \begin{table}[ht!]
% \centering
% \fontsize{8}{9}\selectfont
% \begin{tabular}{cccccc}
% \toprule
% \textbf{Family} & \textbf{Class.} & \textbf{Family} & %\textbf{Class.} & \textbf{Family} & \textbf{Class.} \\
% \midrule
% bigviktor & W & goz & A & bumblebee & A \\
% pizd & W & new\_goz & A & dmsniff & A \\
% ngioweb & W & ccleaner & A & sharkbot & A \\
% verblecon & A & enviserv & A & tufik & A \\
% bazarbackdoor & A & xshellghost & A &  & \\
% \bottomrule
% \end{tabular}
% \caption{Classification of New DGA Families as Word List (W) or %Arithmetic (A)}
% \label{table:generalization}
% \end{table}

%TP- Table \ref{table_Llama3_Pre,F1,Re,NF} presents the detailed evaluation results on the 14 new DGA families showcasing the precision, recall, F1 score, and FPR. %RL- , as well as their category based on its generation scheme (A: arithmetic, or W: word-based). 

Table \ref{table_Llama3_Pre,F1,Re,NF} presents the detailed evaluation results showcasing the precision, recall, and F1 score for the 14 new DGA families. It also shows the FPR obtained.

\begin{table*}[ht!]
\centering
\small
\begin{tabular}{lccc|lccc}
\toprule
\multicolumn{1}{c}{\textbf{Family}} & \textbf{Pre} & \textbf{Re} & \textbf{F1} & \multicolumn{1}{c}{\textbf{Family}} & \textbf{Pre} & \textbf{Re} & \textbf{F1} \\ \midrule
bazarbackdoor  & 0.06 & 0.01 & 0.01 & new\_goz      & 0.96 & 1.00 & 0.98 \\
bigviktor      & 0.88 & 0.35 & 0.50 & ngioweb       & 0.93 & 0.61 & 0.74 \\
bumblebee      & 0.23 & 0.01 & 0.02 & pizd          & 0.95 & 0.90 & 0.93 \\
ccleaner       & 0.83 & 0.22 & 0.35 & sharkbot      & 0.92 & 0.55 & 0.69 \\
dmsniff        & 0.95 & 0.99 & 0.97 & tufik         & 0.96 & 1.00 & 0.98 \\
enviserv       & 0.81 & 0.19 & 0.31 & verblecon     & 0.96 & 1.00 & 0.98 \\
goz            & 0.96 & 1.00 & 0.98 & xshellghost   & 0.96 & 1.00 & 0.98 \\
\bottomrule
\textbf{FPR}                          \textbf{0.05}                           \\
\bottomrule
\end{tabular}
\caption{Metrics obtained with the SFT Llama3 8B model, for new DGA families.}
%TP-\caption{Precision, Recall and F1-Score for New DGA Families for Llama3 8B model.}
\label{table_Llama3_Pre,F1,Re,NF}
\end{table*}

The results revealed a notable performance disparity among the different DGA families. Many generated families, such as \texttt{verblecon}, \texttt{goz}, and \texttt{new\_goz}, demonstrated superior performance with high precision, recall, and F1 score. This suggests that the fine-tuned model exhibits a robust ability to detect and classify these particular types of DGA patterns. It is also important to note that even though the model was challenged with novel normal domains in this test, it exhibited a low FPR.

%TP- In contrast, some families showed significantly lower performance metrics, especially in terms of precision and F1 scores. The family \texttt{bigviktor}, for instance, presented challenges for the model. However, it is important to note that certain generated families, notably \texttt{bumblebee} and \texttt{bazarbackdoor}, also proved difficult for the model to classify accurately, with remarkably low F1 scores. This disparity highlights the challenges associated with detecting certain variants of DGAs, regardless of their generation method, and underscores the need for further refinement of the model to improve its capabilities in these problematic areas

In contrast, some families showed significantly lower performance metrics, such as \texttt{bigviktor}, \texttt{bumblebee}, and \texttt{bazarbackdoor}. These last two are remarkable as they have very low recall and F1 score. This disparity highlights the challenges associated with detecting certain variants of DGAs, regardless of their generation method, and underscores the need for further refinement of the model to improve its capabilities in these problematic areas.

\subsection{Experiment III: Comparison with Previous Approaches}
\label{sec:Comparison with Previous Approaches}

%TP- To evaluate the effectiveness of the LLM-based approach for DGA detection, the performance of the best model, SFT Llama3 8B, is compared against one of the most advanced state-of-the-art models: the LA\_Bin07 model \cite{tuan2022detecting}. The LA\_Bin07 model utilizes a combination of LSTM networks and attention layers to classify domains as either malicious or legitimate.

%TP- To maintain equitable assessment conditions, both the SFT Llama3 8B model and the LA\_Bin07 model underwent evaluation using identical datasets and evaluation procedures. This standardized approach, consistently applied across all models in this study, ensured a fair and comprehensive comparison of their respective performances. 

To evaluate the effectiveness of the LLM-based approach for DGA detection, the SFT Llama3 8B model was compared against one of the most advanced state-of-the-art models: the LA\_Bin07 model \cite{tuan2022detecting}. This model utilizes a combination of LSTM networks and attention layers to classify domains as either malicious or legitimate. 

%TP- To maintain equitable assessment conditions, both the SFT Llama3 8B model and the LA\_Bin07 model were evaluated using identical datasets and evaluation procedures. Tables \ref{tab:comparison} and \ref{tab:comparison_NF_NN} provide a detailed comparison of their performance metrics. Specifically, Table \ref{tab:comparison} refers to the evaluation on the 54 families seen during training, while Table \ref{tab:comparison_NF_NN} focuses on the evaluation of the 14 families not seen during training.

\begin{figure*}[ht!]
    \centering
    \includegraphics[width=\textwidth]{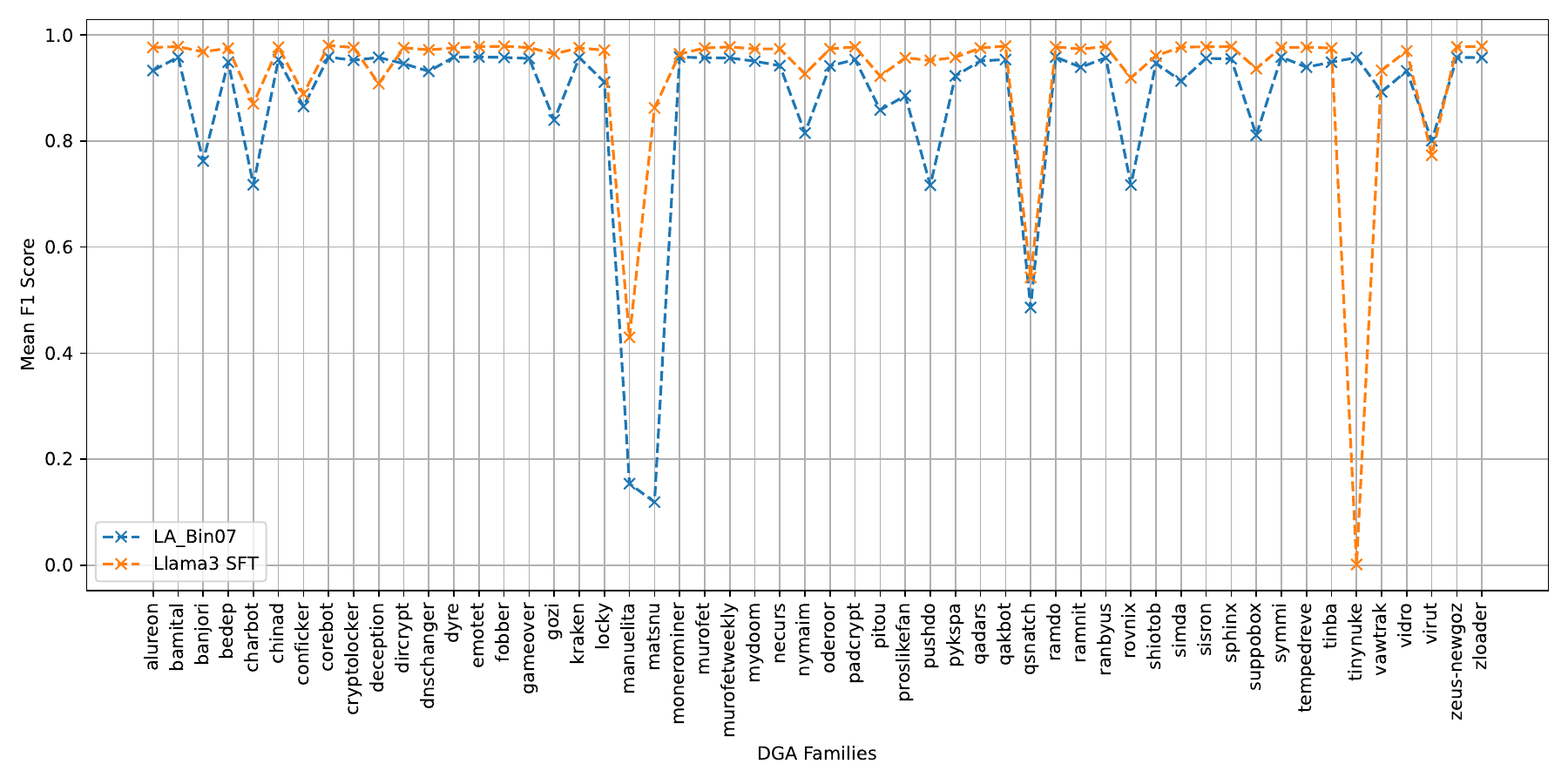}
    \caption{Comparison of mean F1 score between LA\_Bin07 and SFT Llama3 8B  models for 54 DGA families.}
    \label{fig:comparison_metrics_Llama3FT-Labin}
\end{figure*}

\begin{figure*}[ht!]
    \centering
    \includegraphics[width=\textwidth]{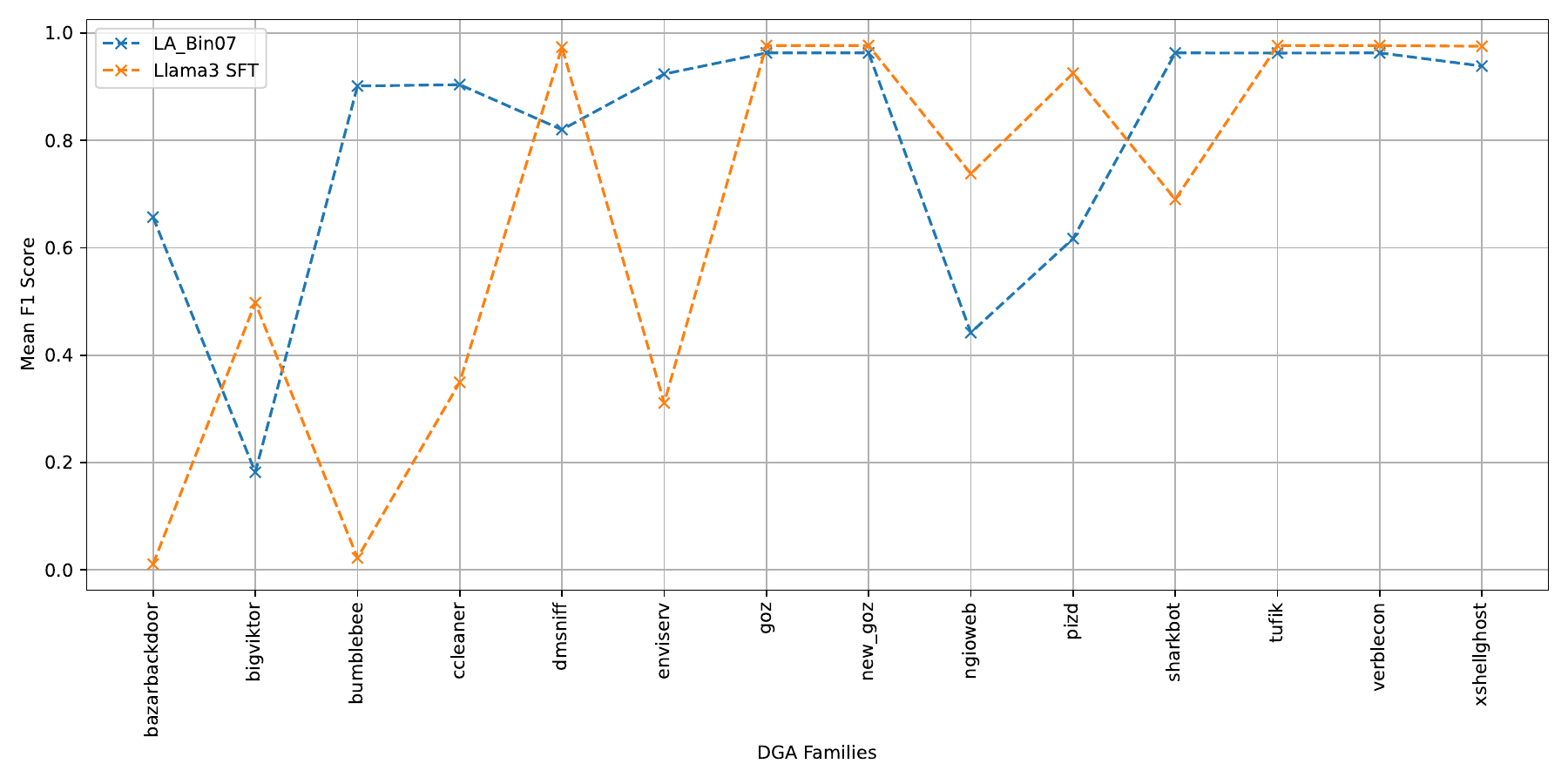}
    \caption{Comparison of mean F1 score between LA\_Bin07 and SFT Llama3 8B models for 14 new DGA families.}
    \label{fig:comparison_metrics_Llama3FT-Labin_NF_NN}
\end{figure*}

To maintain equitable assessment conditions, both the SFT Llama3 8B model and the LA\_Bin07 model were evaluated using identical datasets and evaluation procedures. Tables \ref{tab:comparison} and \ref{tab:comparison_NF_NN} provide a detailed comparison of all their performance metrics, whereas Figures \ref{fig:comparison_metrics_Llama3FT-Labin} and \ref{fig:comparison_metrics_Llama3FT-Labin_NF_NN} highlight the mean F1 score per family. All of these were obtained by evaluating on the previously seen 54 DGA families (Table \ref{tab:comparison} and Figure \ref{fig:comparison_metrics_Llama3FT-Labin}), and the unseen 14 new families (Table \ref{tab:comparison_NF_NN} and Figure \ref{fig:comparison_metrics_Llama3FT-Labin_NF_NN}).

%\subsubsection{Results and Analysis}
%The results of these experiments demonstrated that the fine-tuned Llama3 8B model outperformed the LA\_Bin07 model in several key metrics. Specifically, the model achieved higher accuracy and F1-score while maintaining a lower false positive rate. These results highlight the potential of fine-tuned LLMs in enhancing DGA detection performance, indicating that advanced language models can provide significant advantages over traditional neural network approaches.

\begin{table}[ht]
\centering
\fontsize{8}{9}\selectfont
\begin{tabular}{ccccccc}
\toprule
\textbf{Model} & \textbf{Accu} & \textbf{Pre} & \textbf{Re} & \textbf{F1} & \textbf{FPR} & \textbf{Proc. Time (s)} \\
\midrule
SFT Llama3 8B & 0.94 & 0.93 & 0.92 & 0.92 & 0.04 & 3.50 \\
LA\_Bin07 & 0.90 & 0.90 & 0.88 & 0.88 & 0.09 & 0.03 \\
\bottomrule
\end{tabular}
\caption{Performance comparison between the SFT Llama3 8B model and the LA\_Bin07 model for the previously seen 54 DGA families.}
\label{tab:comparison}
\end{table}

\begin{table}[ht]
\centering
\fontsize{8}{9}\selectfont
\begin{tabular}{ccccccc}
\toprule
\textbf{Model} & \textbf{Accu} & \textbf{Pre} & \textbf{Re} & \textbf{F1} & \textbf{FPR} & \textbf{Proc. Time (s)} \\
\midrule
SFT Llama3 8B &  0.79 & 0.81 & 0.63 & 0.67 & 0.05 & 3.50 \\
LA\_Bin07 &  0.85 & 0.88 & 0.77 & 0.80 & 0.08 & 0.03 \\
\bottomrule
\end{tabular}
\caption{Performance comparison between the SFT Llama3 8B model and the LA\_Bin07 model for the unseen 14 DGA families.}
\label{tab:comparison_NF_NN}
\end{table}
%TP- Table \ref{tab:comparison} provide a detailed comparison of the performance metrics for the SFT Llama3 8B and the LA\_Bin07 models. 

Table \ref{tab:comparison} shows that the SFT Llama3 8B model outperformed the LA\_Bin07 model across several metrics on the previously seen 54 DGA families. Particularly, in the false positive rate (FPR), achieving a value of 4\%, significantly lower than the 9\% observed for LA\_Bin07. Furthermore, Figure \ref{fig:comparison_metrics_Llama3FT-Labin} exhibits that the SFT Llama3 8B model achieves a higher mean F1 score than the LA\_Bin07 model across almost all families, except for \texttt{deception}, \texttt{tyninuke}, and \texttt{virut}. 

Table \ref{tab:comparison} also provides information on processing time. The SFT Llama3 8B model requires 3.50 seconds to process, which is considerably higher than the 0.03 seconds required by the LA\_Bin07 model. This difference highlights a key consideration for real-time applications, where processing speed may be critical. The increased processing time of the SFT Llama3 8B model is likely due to the complexity and size of the LLM, which demands more computational resources while offering superior detection capabilities.

On the other hand, Table \ref{tab:comparison_NF_NN} presents the results obtained by evaluating the SFT Llama3 8B and the LA\_Bin07 models on the 14 DGA families for testing. Although the LA\_Bin07 model demonstrates better performance than Llama3 8B when dealing with these new DGA families, Figure \ref{fig:comparison_metrics_Llama3FT-Labin_NF_NN} reveals a variable performance between the models, with fluctuations in F1 score across different families, indicating that neither model consistently outperforms the other.
%TP- not seen during training. Llama3 8B only outperforms LA\_Bin07 in the FPR metric.

%TP- Figure \ref{fig:comparison_metrics_Llama3FT-Labin} and \ref{fig:comparison_metrics_Llama3FT-Labin_NF_NN} illustrate the F1 score comparison between the LA\_Bin07 and SFT Llama3 8B models for the 54 DGA families seen during training and the 14 families not seen during training. In figure \ref{fig:comparison_metrics_Llama3FT-Labin} it is evident that the SFT Llama3 8B model achieves a higher F1 score than the LA\_Bin07 model across almost all families, except for the \texttt{deception}, \texttt{tyninuke}, and \texttt{virut} families. Conversely, Figure \ref{fig:comparison_metrics_Llama3FT-Labin_NF_NN} reveals a more variable performance between the models, with fluctuations in F1 scores across different cases, indicating that neither model consistently outperforms the other.

\begin{figure*}[ht]
    \centering
    \includegraphics[width=1\textwidth]{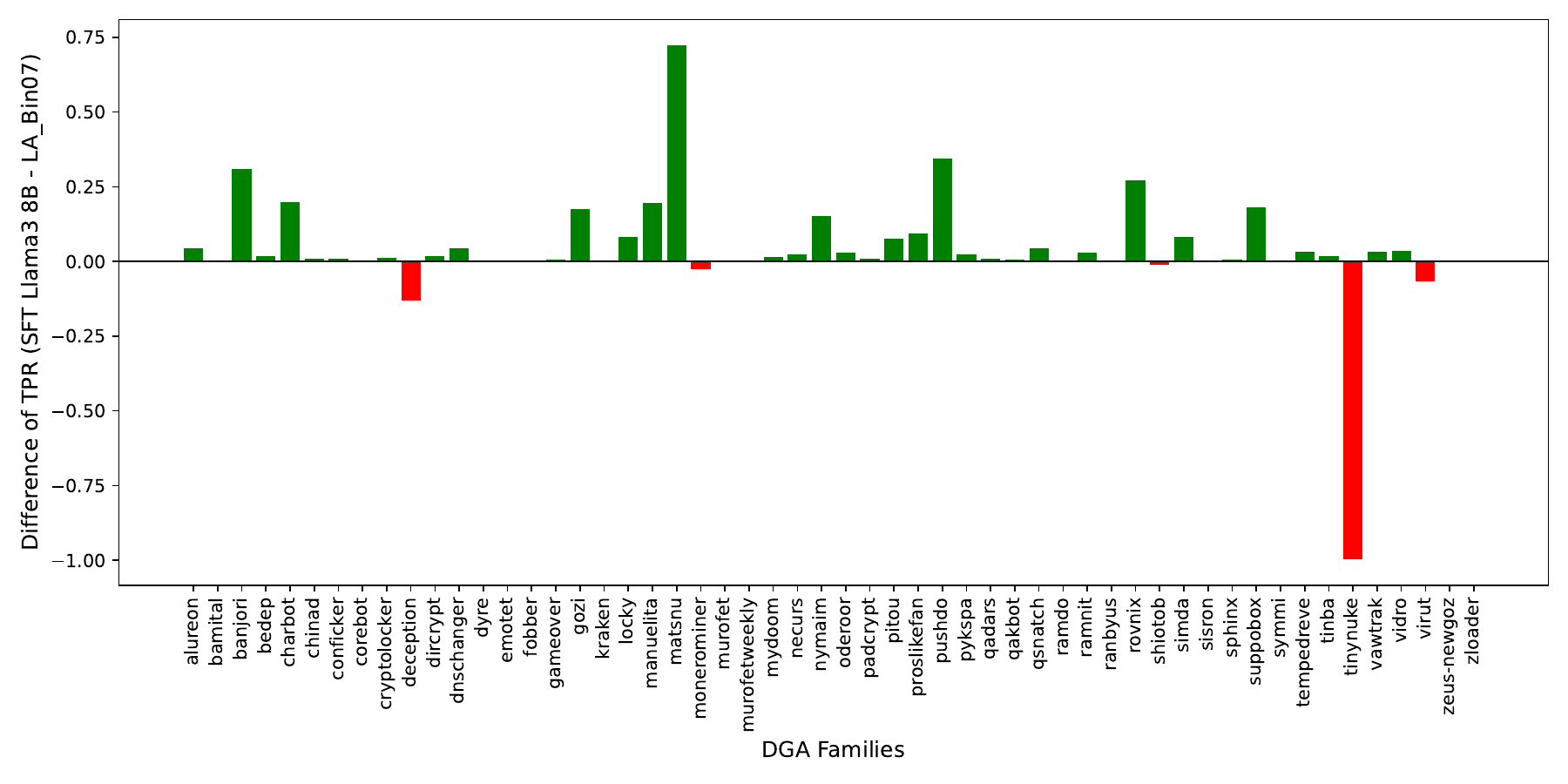}
    \caption{Difference in TPR between SFT Llama3 8B and LA\_Bin07 models for 54 DGA families}
    \label{fig:Difference_TPR}
\end{figure*}

\begin{figure*}[ht]
    \centering
    \includegraphics[width=1\textwidth]{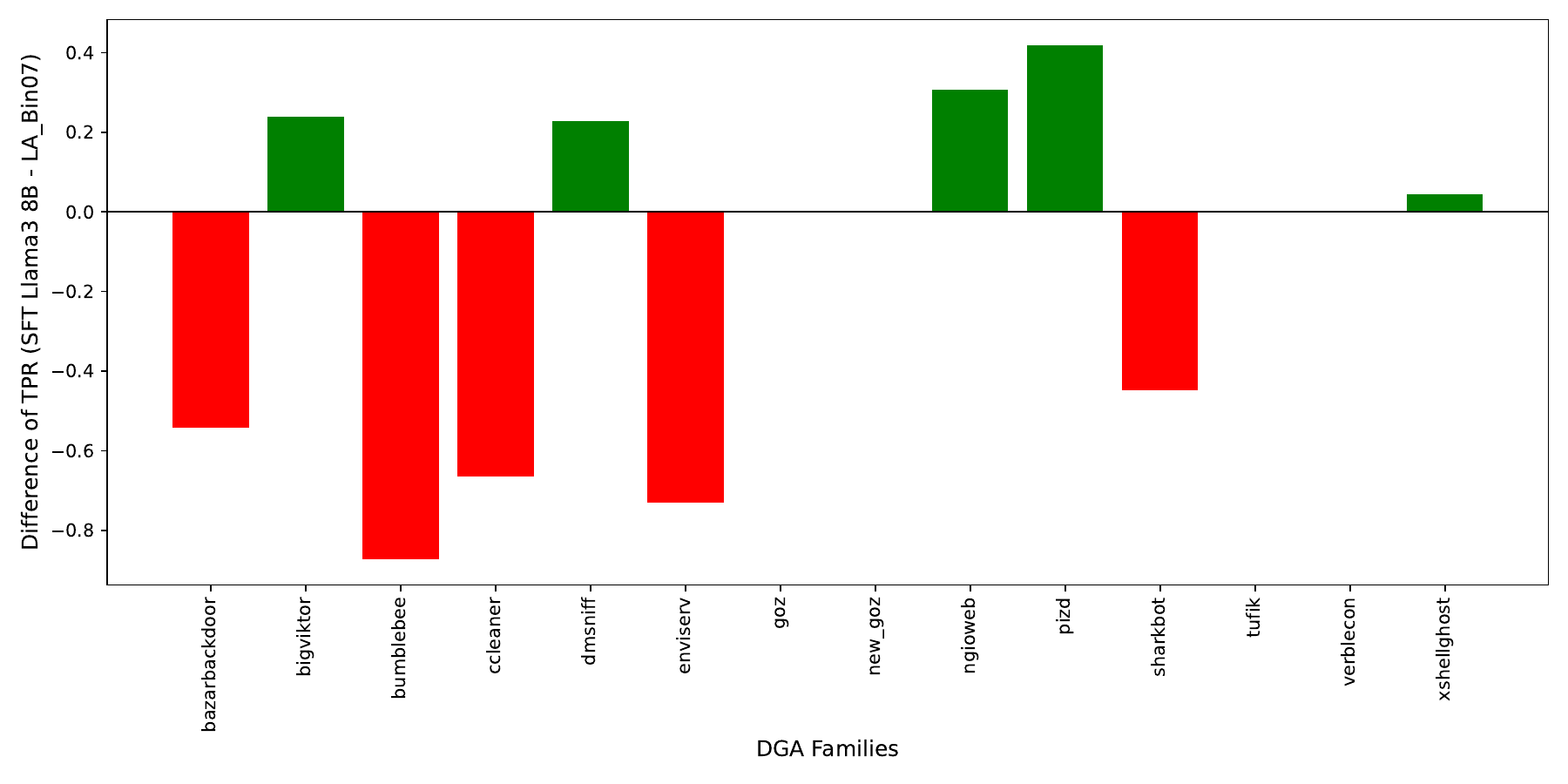}
    \caption{Difference in TPR between SFT Llama3 8B and LA\_Bin07 models for 14 DGA families}
    \label{fig:Difference_TPR_NF_NN}
\end{figure*}

Figures \ref{fig:Difference_TPR} and \ref{fig:Difference_TPR_NF_NN} illustrate the difference in the true positive rate (TPR) between the SFT Llama3 8B model and the LA\_Bin07 model for the 54 DGA families and the 14 new ones, respectively. It can be seen that the first model outperforms the second one in most families present in the training set, which indicates a better capacity to correctly identify malicious domains. Nevertheless, the difference is minor on the new 14 families.  %TP- In Figure \ref{fig:Difference_TPR} it is evident that the first one outperforms the LA\_Bin07 model in most cases, indicating a better capability to correctly identify malicious domains. On the other hand, Figure \ref{fig:Difference_TPR_NF_NN} shows performance fluctuations, with a slightly better performance by the LA\_Bin07 model.%TP- This superiority in TPR highlights the effectiveness of the SFT Llama3 8B model in detecting DGA compared to the LA\_Bin07 model. 

%TP- In this comparative experiment, it was observed that when evaluating the models on the 54 families seen during training, the SFT Llama3 8B model demonstrates substantial improvements in accuracy, precision, recall, F1 score, and false positive rate compared to the LA\_Bin07 model, although with a disadvantage in processing time. However, when the models were evaluated on 14 DGA families not seen during training, a variable performance was observed, with slight improvements by the LA\_Bin07 model compared to the SFT Llama3 8B.

\section{Discussion}
\label{sec:Discution}
%\todo[inline]{we can discuss here, the results, the limitations and possible uses. Here we can also describe the two layered architecture. Very Briefly}

The results from Section~\ref{sec:Experiment} demonstrated the significant potential of LLMs in enhancing DGA detection. In particular, the SFT Llama3 8B model exhibited superior performance to traditional approaches, achieving higher accuracy, precision, recall, and F1 score, while showing the lowest false positive rate.

%Major benefits of LLM were observed in scenarios where conventional methods struggle to keep pace with rapidly evolving threats as shown by the results from the section ~\ref{fig:comparison_metrics_Llama3FT-Labin} where new DGA families were evaluated. In particular, the Llama3 8B SFT  has shown much better results when detecting DGA following a word-based generation algorithm (see Figure~\ref{fig:word-based-results}. The detection Word-based DGA families has been a major issue in DGA detection in the past years~\cite{catania2019analysis}. The current results of LLM shows a promising a approach for dealing with this kind of DGA.

\begin{figure}
    \centering
    \includegraphics[width=1\linewidth]{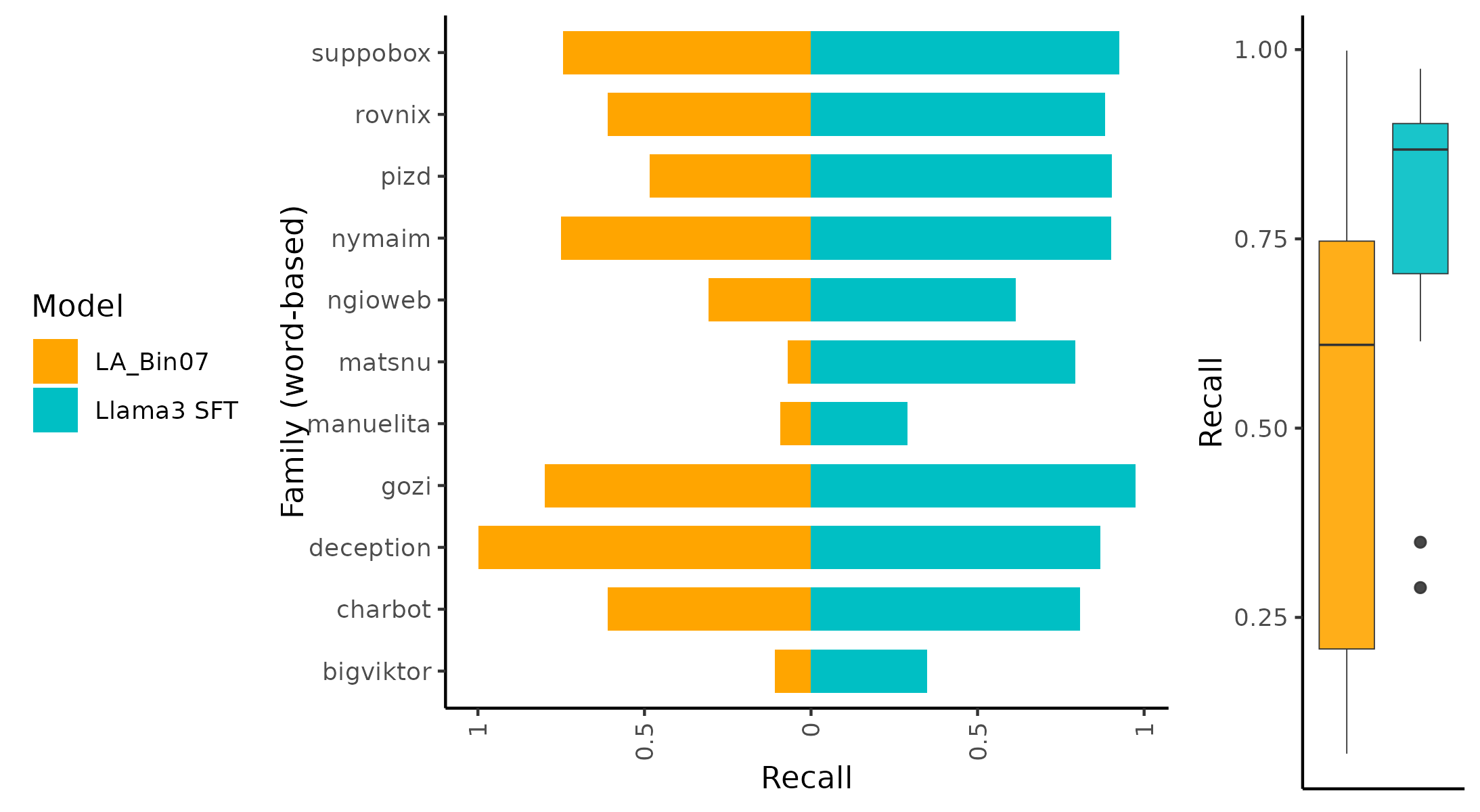}
    \caption{Recall results for LA\_Bin07 and the STF Llama3 8B on word-based DGA families.}
    \label{fig:word-based-results}
\end{figure}

Significant advantages of LLMs were observed in scenarios where conventional methods struggle to keep pace with rapidly evolving threats, as demonstrated by the results in Section \ref{sec:Comparison with Previous Approaches}, where it was compared the SFT Llama3 8B model with the LA\_Bin07 model. Notably, the SFT Llama3 8B model achieved much better results in detecting DGAs that utilize word-based generation algorithms (see Figure~\ref{fig:word-based-results}). Detecting word-based DGA families has been a major challenge in DGA detection in recent years~\cite{catania2019analysis}. The current findings suggest that LLMs offer a promising approach to effectively tackle this type of DGA.

One of the main issues with the application of LLMs for DGA detection is the processing speed of the model. Current LLM responses are a significant limitation for real-time applications, as rapid-response models are required.

To address the processing speed limitations of LLMs in DGA detection, researchers can employ both optimized hardware and smaller, efficient models. Utilizing specialized hardware, such as the Groq platform, which offers high-performance, low-latency processing through optimized matrix multiplication, can significantly enhance model inference and training times. This enables the development of more complex and accurate DGA detection models with reduced computational demands \cite{abts2022groq, moon2024lpu}.

In parallel, adopting smaller yet capable models like Gemma2 2B with 2 billion parameters \cite{team2024gemma} and GPT-4o mini \cite{openai2024} offers a promising approach. These models strike a balance between computational efficiency and detection accuracy, providing a viable alternative for future research \cite{magister2022teaching}.

Finally, a possible approach to address the processing speed limitation could be the application of layered architecture for detection. In the first layer, faster models such as LA\_Bin07 act as initial filters to quickly process domain names. In the next stage, the Llama3 8B model is applied to verify cases identified as suspicious by the primary layer, taking advantage of its high accuracy and low false positive rate.

%This layered approach could allow for the integration of the Llama3 8B model's superior detection capabilities without compromising overall system response time. However, careful calibration will be required to determine the optimal threshold for referral to the secondary layer, balancing processing speed with detection accuracy.

Several promising avenues for future research have been identified. Firstly, exploring continuous learning techniques could keep the model updated with new threats while maintaining performance on known DGAs. Investigating hybrid approaches that combine large language models with other machine learning techniques might optimize both accuracy and efficiency. Additionally, developing methods to enhance model interpretability would enable security analysts to better understand the reasoning behind classifications.  Lastly, conducting  studies to evaluate the model's performance over extended periods of time could assess its resilience to evolving DGA tactics.

\section{Conclusions}
\label{sec:conclusions}

%This study highlights the potential of LLMs in enhancing the detection of DGAs. 

This study on LLMs for DGA detection has proven highly effective, primarily due to their deep understanding of semantics, which enhances DGA word-based detection capabilities. 

Utilizing the Llama3 8B model from Meta, the results demonstrate that both ICL and SFT approaches can significantly improve the identification of DGA domains. Specifically, an SFT  with domain-specific data yields substantial gains in detection accuracy. Furthermore, the combination of low false positive rates and robust detection accuracy makes these models invaluable tools for cybersecurity.

However, challenges remain, particularly in terms of computational power and response time, limiting the feasibility of real-time deployment. To address these issues, exploring smaller models such as Gemma2 2B offers a promising path toward improving efficiency without compromising performance.

Moving forward, further research and development are necessary to seamlessly integrate LLMs into existing network security infrastructures. This includes combining LLMs with other machine learning approaches to create hybrid solutions that optimize both speed and accuracy. Advancing these efforts will be essential to building more effective and scalable defenses against increasingly sophisticated DGA-based threats.

\section*{Acknowledgments}
The first and third authors acknowledge their doctoral fellowship granted by CONICET.

%\clearpage

\bibliographystyle{elsarticle-num} 
\bibliography{cas-refs}

\begin{thebibliography}{10}
\expandafter\ifx\csname url\endcsname\relax
  \def\url#1{\texttt{#1}}\fi
\expandafter\ifx\csname urlprefix\endcsname\relax\def\urlprefix{URL }\fi
\expandafter\ifx\csname href\endcsname\relax
  \def\href#1#2{#2} \def\path#1{#1}\fi

\bibitem{saeed2021survey}
A.~M. Saeed, D.~Wang, H.~A. Alnedhari, K.~Mei, J.~Wang, A survey of machine learning and deep learning based dga detection techniques, in: International Conference on Smart Computing and Communication, Springer, 2021, pp. 133--143.

\bibitem{li2019domain}
S.~Li, T.~Huang, Z.~Qin, F.~Zhang, Y.~Chang, Domain generation algorithms detection through deep neural network and ensemble, in: Companion Proceedings of The 2019 World Wide Web Conference, 2019, pp. 189--196.

\bibitem{zhou2019cnn}
S.~Zhou, L.~Lin, J.~Yuan, F.~Wang, Z.~Ling, J.~Cui, Cnn-based dga detection with high coverage, in: 2019 IEEE international conference on intelligence and security informatics (ISI), IEEE, 2019, pp. 62--67.

\bibitem{hwang2020effective}
C.~Hwang, H.~Kim, H.~Lee, T.~Lee, Effective dga-domain detection and classification with textcnn and additional features, Electronics 9~(7) (2020) 1070.

\bibitem{dombert}
Y.~Tian, Z.~Li, Dom-bert: Detecting malicious domains with pre-training model, in: International Conference on Passive and Active Network Measurement, Springer, 2024, pp. 133--158.

\bibitem{llama3_github}
Meta, \href{https://github.com/meta-llama/llama3}{Llama3: Large language model by meta}, accessed: 2024-07-04 (2024).
\newline\urlprefix\url{https://github.com/meta-llama/llama3}

\bibitem{llama3_huggingface}
Meta, \href{https://huggingface.co/meta-llama/Meta-Llama-3-8B}{Llama3: Large language model by meta}, accessed: 2024-07-04 (2024).
\newline\urlprefix\url{https://huggingface.co/meta-llama/Meta-Llama-3-8B}

\bibitem{plohmann2016comprehensive}
D.~Plohmann, K.~Yakdan, M.~Klatt, J.~Bader, E.~Gerhards-Padilla, A comprehensive measurement study of domain generating malware, in: 25th USENIX Security Symposium (USENIX Security 16), 2016, pp. 263--278.

\bibitem{tranco}
P.~Snyder, C.~Taylor, C.~Kanich, The 2020 tranco list: Improving the alexa ranking, \url{https://tranco-list.eu}, accessed: 2024-07-05 (2020).

\bibitem{devlin2019bert}
J.~Devlin, M.-W. Chang, K.~Lee, K.~Toutanova, Bert: Pre-training of deep bidirectional transformers for language understanding, arXiv preprint arXiv:1810.04805 (2018).

\bibitem{brown2020language}
T.~Brown, B.~Mann, N.~Ryder, M.~Subbiah, J.~D. Kaplan, P.~Dhariwal, A.~Neelakantan, P.~Shyam, G.~Sastry, A.~Askell, et~al., Language models are few-shot learners, Advances in neural information processing systems 33 (2020) 1877--1901.

\bibitem{vaswani2017attention}
A.~Vaswani, N.~Shazeer, N.~Parmar, J.~Uszkoreit, L.~Jones, A.~N. Gomez, {\L}.~Kaiser, I.~Polosukhin, Attention is all you need, Advances in neural information processing systems 30 (2017).

\bibitem{liu2019roberta}
Y.~Liu, M.~Ott, N.~Goyal, J.~Du, M.~Joshi, D.~Chen, O.~Levy, M.~Lewis, L.~Zettlemoyer, V.~Stoyanov, Roberta: A robustly optimized bert pretraining approach, arXiv preprint arXiv:1907.11692 (2019).

\bibitem{wei2022chain}
J.~Wei, X.~Wang, D.~Schuurmans, M.~Bosma, F.~Xia, E.~Chi, Q.~V. Le, D.~Zhou, et~al., Chain-of-thought prompting elicits reasoning in large language models, Advances in neural information processing systems 35 (2022) 24824--24837.

\bibitem{huang2024good}
W.~Huang, X.~Ma, H.~Qin, X.~Zheng, C.~Lv, H.~Chen, J.~Luo, X.~Qi, X.~Liu, M.~Magno, How good are low-bit quantized llama3 models? an empirical study, arXiv preprint arXiv:2404.14047 (2024).

\bibitem{agarwal2024many}
R.~Agarwal, A.~Singh, L.~M. Zhang, B.~Bohnet, S.~Chan, A.~Anand, Z.~Abbas, A.~Nova, J.~D. Co-Reyes, E.~Chu, et~al., Many-shot in-context learning, arXiv preprint arXiv:2404.11018 (2024).

\bibitem{lu2021fantastically}
Y.~Lu, M.~Bartolo, A.~Moore, S.~Riedel, P.~Stenetorp, Fantastically ordered prompts and where to find them: Overcoming few-shot prompt order sensitivity, arXiv preprint arXiv:2104.08786 (2021).

\bibitem{huggingface_sft}
H.~Face, Sfttrainer: Supervised fine-tuning trainer, available at: https://github.com/huggingface/trl [Accessed: 2024-06-15] (2023).

\bibitem{wandb_sft}
W.~. Biases, How to fine-tune an llm part 3: The huggingface trainer, available at: https://wandb.ai/capecape/alpaca\_ft/reports/How-to-Fine-tune-an-LLM-Part-3-The-HuggingFace-Trainer--Vmlldzo5OTEyNjMy [Accessed: 2024-06-15] (2023).

\bibitem{lora_arxiv}
E.~J. Hu, Y.~Shen, P.~Wallis, Z.~Allen-Zhu, Y.~Li, S.~Wang, L.~Wang, W.~Chen, Lora: Low-rank adaptation of large language models, arXiv preprint arXiv:2106.09685 (2021).

\bibitem{lora_github}
Microsoft, Lora: Low-rank adaptation of large language models, available at: https://github.com/microsoft/LoRA [Accessed: 2024-06-15] (2021).

\bibitem{bnb_github}
BitsAndBytes, Bitsandbytes: Optimizers and quantization for large models, available at: https://github.com/TimDettmers/bitsandbytes [Accessed: 2024-06-15] (2022).

\bibitem{yu2018character}
B.~Yu, J.~Pan, J.~Hu, A.~Nascimento, M.~De~Cock, Character level based detection of dga domain names, in: 2018 international joint conference on neural networks (IJCNN), IEEE, 2018, pp. 1--8.

\bibitem{woodbridge2016predicting}
J.~Woodbridge, H.~S. Anderson, A.~Ahuja, D.~Grant, Predicting domain generation algorithms with long short-term memory networks, arXiv preprint arXiv:1611.00791 (2016).

\bibitem{namgung2022efficient}
J.~Namgung, S.~Son, Y.-S. Moon, Efficient deep learning models for dga domain detection, Security and Communication Networks 2021~(1) (2021) 8887881.

\bibitem{catania2019analysis}
C.~Catania, S.~Garc{\'\i}a, P.~Torres, Deep convolutional neural networks for dga detection, in: Computer Science--CACIC 2018: 24th Argentine Congress, Tandil, Argentina, October 8--12, 2018, Revised Selected Papers 24, Springer, 2019, pp. 327--340.

\bibitem{harishkumar2024enhanced}
S.~Harishkumar, R.~Bhuvaneshwaran, Enhanced dga detection in botnet traffic: Leveraging n-gram, topic modeling and attention bilstm (2024).

\bibitem{tapsoba2024analysis}
A.~R. Tapsoba, T.~F. Ou{\'e}draogo, W.-B.~S. Zongo, Analysis of plaintext features in doh traffic for dga domains detection, in: International Conference on Information Technology \& Systems, Springer, 2024, pp. 127--138.

\bibitem{adamwplus}
A.~Javed, I.~Rashid, S.~Tahir, S.~Saeed, A.~M. Almuhaideb, K.~Alissa, Adamw+: Machine learning framework to detect domain generation algorithms for malware, IEEE Access (2024).

\bibitem{alsabeh2024dga}
A.~AlSabeh, K.~Friday, E.~Kfoury, J.~Crichigno, E.~Bou-Harb, On dga detection and classification using p4 programmable switches, Computers \& Security (2024) 104007.

\bibitem{mahdaouy2024domurls_bert}
A.~E. Mahdaouy, S.~Lamsiyah, M.~J. Idrissi, H.~Alami, Z.~Yartaoui, I.~Berrada, Domurls\_bert: Pre-trained bert-based model for malicious domains and urls detection and classification, arXiv preprint arXiv:2409.09143 (2024).

\bibitem{hu2022ci}
H.~Wang, Z.~Tang, H.~Li, J.~Zhang, S.~Li, J.~Wang, Ci\_gru: An efficient dga botnet classification model based on an attention recurrence plot, Computer Networks 235 (2023) 109992.

\bibitem{tuan2022detecting}
T.~A. Tuan, H.~V. Long, D.~Taniar, On detecting and classifying dga botnets and their families, Computers \& Security 113 (2022) 102549.

\bibitem{cebere2024down}
B.~Cebere, J.~Flueren, S.~Sebasti{\'a}n, D.~Plohmann, C.~Rossow, Down to earth! guidelines for dga-based malware detection (2024).

\bibitem{zago2020umudga}
M.~Zago, M.~G. P{\'e}rez, G.~M. P{\'e}rez, Umudga: A dataset for profiling algorithmically generated domain names in botnet detection, Data in Brief 30 (2020) 105400.

\bibitem{360netlab}
360NetLab, \href{https://data.netlab.360.com/}{360netlab dga dataset}, accessed: 2024-07-04 (2023).
\newline\urlprefix\url{https://data.netlab.360.com/}

\bibitem{dga_repo}
J.~Bader, \href{https://github.com/baderj/domain\_generation\_algorithms}{Domain generation algorithms repository}, accessed: 2024-07-04 (2024).
\newline\urlprefix\url{https://github.com/baderj/domain\_generation\_algorithms}

\bibitem{yao2022react}
S.~Yao, J.~Zhao, D.~Yu, N.~Du, I.~Shafran, K.~Narasimhan, Y.~Cao, React: Synergizing reasoning and acting in language models, arXiv preprint arXiv:2210.03629 (2022).

\bibitem{plohmann2015dgaarchive}
D.~Plohmann, Dgaarchive--a deep dive into domain generating malware, Dec-2015 (2015).

\bibitem{peck2019charbot}
J.~Peck, C.~Nie, R.~Sivaguru, C.~Grumer, F.~Olumofin, B.~Yu, A.~Nascimento, M.~De~Cock, Charbot: A simple and effective method for evading dga classifiers, IEEE Access 7 (2019) 91759--91771.

\bibitem{ollama}
Ollama, Ollama: An open source package for optimizing language models, available at: https://github.com/Ollama/ollama [Accessed: 2024-06-15] (2023).

\bibitem{dettmers2024qlora}
T.~Dettmers, A.~Pagnoni, A.~Holtzman, L.~Zettlemoyer, Qlora: Efficient finetuning of quantized llms, Advances in Neural Information Processing Systems 36 (2024).

\bibitem{cochran1977sampling}
W.~G. Cochran, Sampling Techniques, 3rd Edition, John Wiley \& Sons, New York, 1977.

\bibitem{hossin2015review}
M.~Hossin, M.~N. Sulaiman, A review on evaluation metrics for data classification evaluations, International journal of data mining \& knowledge management process 5~(2) (2015) 1.

\bibitem{powers2020evaluation}
D.~M. Powers, Evaluation: from precision, recall and f-measure to roc, informedness, markedness and correlation, arXiv preprint arXiv:2010.16061 (2020).

\bibitem{reypapin2024dga}
R.~Leyva, Domain-name-classification-with-llm, \url{https://github.com/reypapin/Domain-Name-Classification-with-LLM} (2024).

\bibitem{abts2022groq}
D.~Abts, J.~Kim, G.~Kimmell, M.~Boyd, K.~Kang, S.~Parmar, A.~Ling, A.~Bitar, I.~Ahmed, J.~Ross, The groq software-defined scale-out tensor streaming multiprocessor: From chips-to-systems architectural overview, in: 2022 IEEE Hot Chips 34 Symposium (HCS), IEEE Computer Society, 2022, pp. 1--69.

\bibitem{moon2024lpu}
S.~Moon, J.-H. Kim, J.~Kim, S.~Hong, J.~Cha, M.~Kim, S.~Lim, G.~Choi, D.~Seo, J.~Kim, et~al., Lpu: A latency-optimized and highly scalable processor for large language model inference, IEEE Micro (2024).

\bibitem{team2024gemma}
G.~Team, M.~Riviere, S.~Pathak, P.~G. Sessa, C.~Hardin, S.~Bhupatiraju, L.~Hussenot, T.~Mesnard, B.~Shahriari, A.~Ram{\'e}, et~al., Gemma 2: Improving open language models at a practical size, arXiv preprint arXiv:2408.00118 (2024).

\bibitem{openai2024}
OpenAI, \href{https://openai.com/index/gpt-4o-mini-advancing-cost-efficient-intelligence/}{Gpt-4o mini: Advancing cost-efficient intelligence}, accessed: 2024-08-05 (2024).
\newline\urlprefix\url{https://openai.com/index/gpt-4o-mini-advancing-cost-efficient-intelligence/}

\bibitem{magister2022teaching}
L.~C. Magister, J.~Mallinson, J.~Adamek, E.~Malmi, A.~Severyn, Teaching small language models to reason, arXiv preprint arXiv:2212.08410 (2022).

\end{thebibliography}

\end{document}